# The Role of Macros in Tractable Planning

**Anders Jonsson**                                    ANDERS.JONSSON@UPF.EDU
*Dept. of Information and Communication Technologies*
*Universitat Pompeu Fabra*
*Roc Boronat, 138*
*08018 Barcelona, Spain*

## Abstract

This paper presents several new tractability results for planning based on macros. We describe an algorithm that optimally solves planning problems in a class that we call inverted tree reducible, and is provably tractable for several subclasses of this class. By using macros to store partial plans that recur frequently in the solution, the algorithm is polynomial in time and space even for exponentially long plans. We generalize the inverted tree reducible class in several ways and describe modifications of the algorithm to deal with these new classes. Theoretical results are validated in experiments.

## 1. Introduction

A growing number of researchers in planning are investigating the computational complexity of solving different classes of planning problems. Apart from the theoretical value of such knowledge, of particular interest are tractable planning problems that can be provably solved in polynomial time. Many, if not most, state-of-the-art planners are based on heuristic search. Acquiring an informative heuristic requires quickly solving a relaxed version of the planning problem. Known tractable classes of planning problems are ideal candidates for projecting the original problem onto in order to generate heuristics.

This paper introduces the class **IR** of inverted tree reducible planning problems and presents an algorithm that uses macros to solve instances of this class. The algorithm is provably complete and optimal for **IR**, and its complexity depends on the size of the domain transition graph that the algorithm constructs. In the worst case, the size of the graph is exponential in the number of state variables, but we show that the algorithm runs in polynomial time for several subclasses of **IR**, proving that plan generation is tractable for these classes.

The tractable subclasses of **IR** include planning problems whose optimal solution has exponential length in the number of state variables. The reason that our algorithm can solve these problems in polynomial time is that subsequences of operators are frequently repeated in the solution. Since the algorithm stores each such subsequence as a macro, it never needs to generate the subsequence again. The algorithm can be viewed as a compilation scheme where state variables are compiled away. The only information kept about each state variable is the set of macros acting on it.

We extend the class **IR** in two ways and modify our algorithm so that it solves planning problems in the new classes. First, we define the notion of a relaxed causal graph, which contains less edges than the conventional causal graph, and an associated class **RIR** of relaxed inverted tree reducible planning problems. We also define the notion of reversible variables, and define the class **AR** of planning problems with acyclic causal graphs and reversible variables. We show that our algorithm can be modified to solve planning problems in this class. Finally, we combine the results for **IR** and **AR** to define the class **AOR** of planning problems with acyclic causal graphs and only some of the variables reversible.

The class **IR** and the algorithm for solving instances of the class previously appeared in a conference paper (Jonsson, 2007). The present paper includes new tractability results for the classes **IR**, **AR**, and **AOR**. In addition, the notion of relaxed causal graphs and the resulting class **RIR** are novel.

Perhaps the most important contribution of this work is establishing several novel tractability results for planning. In particular, we present several classes of planning problems that can be solved in polynomial time. This contribution is particularly significant since the number of known tractable classes of planning problems is very small. In addition, some of our new classes can be solved optimally. The novel approach taken by the paper is the use of macros to store solutions to subproblems, making it possible to represent exponentially long plans in polynomial time and space.

A related contribution is the possibility to use the tractable classes to generate heuristics. We discuss two possibilities for generating heuristics here. One idea is to project a planning problem onto **IR** by removing some pre-conditions and effects of operators to make the causal graph acyclic and inverted tree reducible. In fact, this is the strategy used by Helmert (2006) to compute the causal graph heuristic. Next, we could duplicate each operator by introducing pre-conditions on "missing" variables. The resulting problem is tractable according to Theorem 3.5. Thus, the algorithm could solve the projected problem in polynomial time to compute an admissible heuristic for the original problem. Note, however, that the number of resulting actions is exponential in the number of missing variables, so the strategy will only be tractable when the number of missing variables is a small constant.

Another idea is to perform the same reduction until the causal graph is acyclic and each variable has a constant number of ancestors. Next, we could make the current problem part of the class **AOR** by introducing actions to make certain variables reversible. Both types of changes have the effect of making the problem easier to solve, in the sense that a solution to the relaxed problem is guaranteed to be no longer than the solution to the original problem. We could then solve the resulting problem using our algorithm for **AOR**. Although this algorithm is not provably optimal, it might be an informative heuristic in some cases.

Another contribution is the possibility of solving some planning problems in polynomial time, since they fall into one of the classes of planning problems studied in this paper. For example, we show that this is possible for the Gripper and Logistics domains. Although these domains were previously known to be polynomial-time solvable, it was not known whether they could be solved using the type of macro approach taken here. If a planning problem cannot be solved directly, parts of the causal graph might have the structure required by our algorithms. In this case, the algorithms could be used to solve part of the planning problem. Since one feature of the algorithms is compiling away variables and replacing them with macros, the reduced problem could then be fed to a standard planner to obtain the final solution.

Perhaps a more interesting use of the algorithms is the possibility to reuse macros. If part of the causal graph structure coincides between two problems, macros generated in one problem could immediately be substituted in the other problem. Since variables are compiled away and replaced by macros, the resulting problem has fewer variables and actions. For example, in Tower of Hanoi, the macros generated for the $n-1$ disc problem could immediately be substituted in the $n$ disc problem. The resulting problem only includes the macros for the $n-1$ disc problem and a single variable corresponding to the $n$-th disc.

The rest of the paper is organized as follows. Section 2 introduces notation and definitions. Section 3 introduces the class **IR** of inverted tree reducible planning problems, presents an algorithm for solving problems in this class, and proves several theoretical properties about the algorithm. Section 4 introduces several extensions of the class **IR**, as well as corresponding algorithms. Section 5 presents experimental results validating the theoretical properties of the algorithms. Section 6 relates the paper to existing work in planning, and Section 7 concludes with a discussion.

## 2. Notation

Let $V$ be a set of variables, and let $D(v)$ be the finite domain of variable $v \in V$. We define a state $s$ as a function on $V$ that maps each variable $v \in V$ to a value $s(v) \in D(v)$ in its domain. A partial state $x$ is a function on a subset $V_x \subseteq V$ of variables that maps each variable $v \in V_x$ to $x(v) \in D(v)$. Sometimes we use the notation $(v_1 = x_1, \ldots, v_k = x_k)$ to denote a partial state $x$ defined by $V_x = \{v_1, \ldots, v_k\}$ and $x(v_i) = x_i$ for each $v_i \in V_x$.

We define several operations on partial states. For a subset $W \subseteq V$ of variables, $x \mid W$ is the partial state obtained by restricting the scope of $x$ to $V_x \cap W$. Two partial states $x$ and $y$ *match*, which we denote $x \sim y$, if $x \mid V_y = y \mid V_x$, i.e., $x(v) = y(v)$ for each $v \in V_x \cap V_y$. Given two partial states $x$ and $y$, let $x \oplus y$, the *composition* of $x$ and $y$, be a partial state $z$ defined by $V_z = V_x \cup V_y$, $z(v) = y(v)$ for each $v \in V_y$, and $z(v) = x(v)$ for each $v \in V_x - V_y$. Note that composition is not symmetric since the right operand overrides the values of the left operand.

A planning problem is a tuple $P = \langle V, init, goal, A \rangle$, where $V$ is the set of variables, $init$ is an initial state, $goal$ is a partial goal state, and $A$ is a set of actions. An action $a = \langle pre(a); post(a) \rangle \in A$ consists of a partial state $pre(a)$ called the *pre-condition* and a partial state $post(a)$ called the *post-condition*. Action $a$ is applicable in any state $s$ such that $s \sim pre(a)$ and results in a new state $s \oplus post(a)$.

The *causal graph* of a planning problem $P$ is a directed graph $(V, E)$ with the variables as nodes. There is an edge $(w, v) \in E$ if and only if $w \neq v$ and there exists an action $a \in A$ such that $w \in V_{pre(a)} \cup V_{post(a)}$ and $v \in V_{post(a)}$. If the causal graph is acyclic, each action $a \in A$ is unary, i.e., $|V_{post(a)}| = 1$. For each variable $v \in V$, we define $Anc(v)$ as the set of ancestors of $v$ in the causal graph, and $Desc(v)$ as the set of descendents of $v$.

A macro-action, or macro for short, is an ordered sequence of actions. For a macro to be well-defined, the pre-condition of each action has to coincide with the cumulative post-condition of the actions that precede it in the sequence. Although an action sequence implicitly induces a pre- and a post-condition, we explicitly associate a pre- and a post-condition with each macro. Since a macro is functionally an action, we can define hierarchies of macros such that the action sequence of a macro includes other macros, as long as this does not cause the definitions of macros to be cyclic.

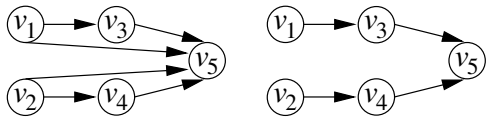

Figure 1: An acyclic causal graph and its transitive reduction.

**Definition 2.1.** *A sequence of actions $seq = \langle a_1, \ldots, a_k \rangle$ is well-defined for a state $s$ if $s \sim pre(a_1)$ and $s \oplus post(a_1) \oplus \ldots \oplus post(a_{i-1}) \sim pre(a_i)$ for each $i \in \{1, \ldots, k\}$.*

We sometimes refer to the post-condition $post(seq)$ of an action sequence, defined as $post(seq) = post(a_1) \oplus \ldots \oplus post(a_k)$. Given two sequences $seq1$ and $seq2$, $\langle seq1, seq2 \rangle$ is the concatenation of the two sequences, and $post(\langle seq1, seq2 \rangle)$ is the post-condition of the resulting sequence, defined as $post(seq1) \oplus post(seq2)$. We are now ready to formally define macros as we use them in the paper:

**Definition 2.2.** *A macro $m = \langle pre(m), seq(m), post(m) \rangle$ consists of a pre-condition $pre(m)$, a sequence $seq(m) = \langle a_1, \ldots, a_k \rangle$ of actions in $A$ and other macros, and a post-condition $post(m)$. The macro $m$ is well-defined if $seq(m)$ is well-defined for $pre(m)$ and $pre(m) \oplus post(seq(m)) = post(m)$.*

## 3. The Class IR

Katz and Domshlak (2008a) defined **I** as the class of planning problems whose causal graphs are *inverted trees*, i.e., the outdegree of each variable in the causal graph is less than or equal to 1, and there is a unique root variable $v$ with outdegree 0. In this section, we study a class of planning problems that we call **IR**, which stands for *inverted tree reducible*:

**Definition 3.1.** *A planning problem $P$ is inverted tree reducible if and only if the causal graph of $P$ is acyclic and the transitive reduction of the causal graph is an inverted tree.*

The transitive reduction $(V, E')$ of a graph $(V, E)$ is defined on the same set of nodes $V$, while $E'$ contains the minimal set of edges such that the transitive closure of $E'$ is equal to the transitive closure of $E$. In other words, the transitive reduction only retains edges necessary to maintain connectivity. For acyclic graphs, the transitive reduction is unique and can be computed efficiently. Figure 1 illustrates the causal graph and its transitive reduction for a planning problem in the class **IR**. Here, the root variable is $v_5$.

For a graph $G = (V, E)$ on the variables $V$ of a planning problem $P$ and each variable $v \in V$, let $Pa(v) = \{w \in V : (w, v) \in E\}$ be the set of parents of $v$ in $G$, i.e., variables on incoming edges. In this paper, we always refer to the parents of $v$ in the transitive reduction $(V, E')$, as opposed to the causal graph itself. For example, in Figure 1 the parents of $v_5$ in the causal graph are $\{v_1, \ldots, v_4\}$, but $Pa(v_5) = \{v_3, v_4\}$. Note that the transitive reduction preserves the set of ancestors $Anc(v)$ and descendants $Desc(v)$ of a variable $v$.

Without loss of generality, in what follows we assume that the goal state is defined on the root variable $v$ of the transitive reduction, i.e., $v \in V_{goal}$. Since there are no outgoing edges from $v$, the pre-condition of any action $a \in A$ that changes the value of a variable $w \neq v$ is independent of $v$. If $v \notin V_{goal}$, a plan that solves $P$ never needs to change the value of $v$, so we can eliminate $v$ from the problem, resulting in one or several subgraphs of the causal graph. It is trivial to show that the transitive reductions of the resulting subgraphs are inverted trees, so the problem can be decomposed into one or several planning problems in **IR** that can be solved independently.

For convenience, we add a dummy variable $v^*$ to $V$, as well as a dummy action $a^* = \langle goal; (v^* = 1) \rangle$ to $A$. The transitive reduction of the resulting causal graph contains an additional node $v^*$ and an additional edge $(v, v^*)$, where $v$ is the original root variable (note that $v^*$ becomes the root variable of the new transitive reduction). Since $v \in V_{goal}$ by assumption, the causal graph contains the edge $(v, v^*)$, and $v^*$ is reachable from each other variable in the transitive reduction via $v$.

The rest of Section 3 is organized as follows. Subsection 3.1 presents MacroPlanner, an algorithm that solves planning problems in the class **IR** via decomposition into subproblems. The subsection includes detailed descriptions of how the subproblems are generated and solved. Subsection 3.2 shows examples of how MacroPlanner solves planning problems in **IR**. Subsection 3.3 proves several theoretical properties of MacroPlanner, and Subsection 3.4 discusses how the algorithm can be improved by pruning the search for macros.

## 3.1 Plan Generation for IR

In this section, we present a plan generation algorithm for the class **IR** called MacroPlanner. For each planning problem $P \in$ **IR**, MacroPlanner generates one or several plans that solve $P$ in the form of macros. The algorithm uses a divide-and-conquer strategy to define and solve several subproblems for each variable $v \in V$ of the problem. It then stores the solutions to each subproblem as macros, and incorporates these macros into the action sets of higher-level subproblems.

---

**Algorithm 1** MacroPlanner($P$)

---
1: $G \leftarrow$ causal graph of $P$
2: $R \leftarrow$ transitive reduction of $G$
3: $v \leftarrow$ root variable of $R$
4: $M \leftarrow$ GetMacros($v, init, A, R$)
5: **return** $M$, or "fail" if $M = \emptyset$

---

The main routine of MacroPlanner appears in Algorithm 1. MacroPlanner takes a planning problem $P \in$ **IR** as input and constructs the causal graph $G$ of $P$, as well as the transitive reduction $R$ of $G$. It then identifies the root variable $v$ in the transitive reduction, and calls GetMacros($v, init, A, R$) to solve $P$. Note that $M$ is a set of macros, implying that MacroPlanner may return multiple solutions to $P$. If $M$ does not contain any solution to $P$, MacroPlanner returns "fail". Also note that $v$ refers to the original root variable of the transitive reduction (excluding the dummy variable $v^*$). However, in what follows we still consider $v^*$ a descendant of $v$.

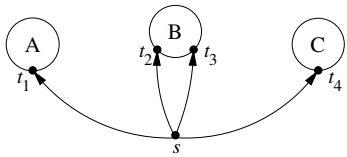

Figure 2: Illustration of the macros generated by MacroPlanner.

### 3.1.1 Defining Subproblems

In this section we describe the subroutine GetMacros which, for each variable $v \in V$, defines several subproblems for $v$. Intuitively, there are only two reasons to change the values of $v$ and its ancestors: to satisfy the pre-condition of some action $a \in A$, or to reach the goal state. The idea is to exhaustively define all subproblems for $v$ that may be needed to achieve this.

Let $P_v = \langle V_v, init', goal', O \rangle$ be a subproblem associated with variable $v \in V$. The set of variables $V_v = \{v\} \cup Anc(v)$ consists of $v$ and its ancestors in the causal graph. The set of actions $O$ contains each action that changes the value of $v$ and each macro representing the solution to a subproblem for a parent $w \in Pa(v)$ of $v$. Since $V_v$ and $O$ are fixed, two subproblems for $v$ only differ in their initial and goal states $init'$ and $goal'$.

To define all subproblems that may be needed to change the values of $v$ and its ancestors, we project onto $V_v$ the pre-condition of each action that changes the value of some descendant of $v$. Let $Z = \{pre(a) \mid V_v : a \in A \wedge V_{post(a)} \subseteq Desc(v)\} - \{()\}$ be the set of such projected pre-conditions. We exclude the empty partial state () from $Z$, implying $|V_z| > 0$ for each $z \in Z$. Note that $Z$ includes the projected goal state $goal \mid V_v$ since $goal$ is a pre-condition of the dummy action $a^*$ that changes the value of $v^*$, a descendant of $v$.

For each projected pre-condition $z \in Z$, GetMacros defines a subproblem $P_v$ with initial state $init' = init \mid V_v$ and goal state $goal' = z$. GetMacros calls the subroutine Solve to generate one or several solutions to $P_v$, each in the form of a sequence $seq = \langle a_1, \ldots, a_k \rangle$ of actions and macros in $O$. Let $s = init' \oplus post(seq)$ be the state that results from applying $seq$ in $init'$. Then Solve returns the solution to $P_v$ in the form of a macro $m = \langle init', seq, s \rangle$, where $init'$ and $s$ are fully specified states for $V_v$. For each such macro $m$ and each $z \in Z$, GetMacros defines a new subproblem $P_v$ with initial state $init' = post(m)$ and goal state $goal' = z$. This process continues until no new subproblems are defined.

The subroutine Solve may generate multiple macros for satisfying the same projected pre-condition. An intuitive understanding of what macros are generated is provided by Lemma B.3. The lemma states that for each state $u$ that is reachable from $init'$ and each projected pre-condition $z \in Z$ such that $u \sim z$, Solve generates a macro $\langle init', seq, t \rangle$ such that $t \sim z$ and $t$ is on a shortest path from $init'$ to $u$ with prefix $seq$. In other words, Solve generates only the shortest possible macros for satisfying $z$. However, it might have to generate multiple such macros to ensure completeness and optimality.

Figure 2 illustrates the macros that MacroPlanner generates. Let $s$ be the current state, let $z$ be a projected pre-condition, and let $A$, $B$, and $C$ be three sets of states that satisfy $z$. If $t_1$ is on a shortest path from $s$ to any state in $A$, MacroPlanner generates a macro from $s$ to $t_1$ that represents such a shortest path (the same holds for $t_4$ and $C$). If no state in $B$ is on the shortest path from $s$ to all other states in $B$, the algorithm generates several macros to states in $B$, in this case to $t_2$ and $t_3$, such that either $t_2$ or $t_3$ is on a shortest path from $s$ to any state in $B$.

---

**Algorithm 2** GetMacros($v, init, A, G$)

---

1:   $M \leftarrow \emptyset$
2:   $L \leftarrow$ list containing the projected initial state $init \mid V_v$
3:   $Z \leftarrow \{pre(a) \mid V_v : a \in A \wedge V_{post(a)} \subseteq Desc(v)\} - \{()\}$
4:   $O \leftarrow \{a \in A : v \in V_{post(a)}\}$
5:   **for all** $w \in Pa(v)$ **do**
6:     $O \leftarrow O \cup$ GetMacros($w, init, A, G$)
7:   **end for**
8:   **while** there are more elements in $L$ **do**
9:     $s \leftarrow$ next element in $L$
10:    $M \leftarrow M \cup$ Solve($v, s, Z, O, G$)
11:    **for all** $m = \langle pre(m), seq(m), post(m) \rangle \in M$ such that $post(m) \notin L$ **do**
12:      append $post(m)$ to $L$
13:    **end for**
14:   **end while**
15:   **return**   $M$

---

Algorithm 2 gives pseudo-code for GetMacros. The input of GetMacros is a variable $v \in V$, an initial state $init$, an action set $A$, and a graph $G = (V, E)$ on the variables $V$. To obtain the action set $O$, GetMacros recursively calls itself for each parent $w \in Pa(v)$ of $v$. The list $L$ is initialized with the projected initial state $init \mid V_v$. The subroutine Solve, which is described in the following section, simultaneously solves all subproblems for $v$ with initial state $s$ and returns the corresponding macros. For each distinct post-condition $post(m)$ of a macro $m$ returned by Solve, GetMacros adds $post(m)$ to the list $L$. Consequently, Solve is later called with initial state $post(m)$. Finally, GetMacros returns all macros generated by Solve.

### 3.1.2 Solving the Subproblems

In this section we describe the subroutine Solve, which solves the subproblems defined by GetMacros for a variable $v \in V$. The idea is to construct a graph with the states for $V_v$ as nodes. This graph can be seen as a joint domain transition graph of the variables in $V_v$. For each initial state $s$ and each projected pre-condition $z \in Z$, Solve computes the shortest path from $s$ to states that match $z$ using a straightforward application of Dijkstra's algorithm. However, the number of states is exponential in the cardinality of $V_v$, so Solve constructs the graph implicitly, only adding nodes as needed.

For each state $p$ in the graph, Solve generates two types of successor states. Successor states of the first type are those that change the value of $v$. To find such states, Solve finds each action $a \in O$ such that $pre(a) \sim (v = p(v))$, i.e., the pre-condition of $a$ is satisfied with respect to the current value of $v$. For each parent $w \in Pa(v)$ such that $(p \mid V_w) \not\sim (pre(a) \mid V_w)$, Solve looks for macros in $O$ from $p \mid V_w$ to states that satisfy $pre(a) \mid V_w$. For each sequence $seq$ that represents a combination of such macros, Solve adds a successor state $p \oplus post(\langle seq, a \rangle)$ of $p$ to the graph. The cost of the corresponding edge is the length of the sequence $\langle seq, a \rangle$.

Successor states of the second type are those that match a projected pre-condition $z \in Z$. To find such states, Solve finds each $z \in Z$ such that $z \sim (v = p(v))$. For each parent $w \in Pa(v)$ such that $(p \mid V_w) \not\sim (z \mid V_w)$, Solve looks for macros in $O$ from $p \mid V_w$ to states that match $z \mid V_w$. For each sequence $seq$ that represents a combination of such macros, Solve generates a successor state $p \oplus post(seq)$ of $p$. However, this successor state is not added to the graph. Instead, it is only used to generate a solution to the corresponding subproblem by defining a macro from the initial state $s$ to the state $p \oplus post(seq)$.

---

**Algorithm 3** Solve$(v, s, Z, O, G)$

1: $M \leftarrow \emptyset$
2: $Q \leftarrow$ priority queue containing the pair $(s, \langle \rangle)$
3: **while** $Q$ is non-empty **do**
4:     $(p, seq) \leftarrow$ remove highest priority state-sequence pair from $Q$
5:     **for all** $a \in O$ such that $pre(a) \sim (v = p(v))$ **do**
6:         $S \leftarrow$ Compose$(v, p, pre(a), O, G)$
7:         **for all** action sequences $seq2 \in S$ **do**
8:             insert $(p \oplus post(\langle seq2, a \rangle), \langle seq, seq2, a \rangle)$ into $Q$
9:         **end for**
10:     **end for**
11:     **for all** $z \in Z$ such that $z \sim (v = p(v))$ **do**
12:         $S \leftarrow$ Compose$(v, p, z, O, G)$
13:         **for all** action sequences $seq2 \in S$ **do**
14:             $M \leftarrow M \cup \{\langle s, \langle seq, seq2 \rangle, p \oplus post(seq2) \rangle\}$
15:         **end for**
16:     **end for**
17: **end while**
18: **return** $M$

---

Algorithm 3 gives pseudo-code for Solve, which takes as input a variable $v \in V$, an initial state $s$, a set of projected pre-conditions $Z$, an action set $O$, and a graph $G = (V, E)$ on $V$. The priority queue $Q$ contains state-sequence pairs, and elements are ordered according to the total length of the associated sequence. Successor states of the first type are generated on lines 5–10, and successor states of the second type are generated on lines 11–16. Inserting a state-sequence pair $(p, seq)$ into $Q$ (line 8) only succeeds if $seq$ is the shortest sequence to $p$ so far, replacing any previous state-sequence pair involving $p$. Likewise, inserting a macro $\langle s, seq, p \rangle$ into $M$ (line 14) only succeeds if $seq$ is the shortest sequence to $p$ so far, replacing any previous macro to $p$.

Algorithm 4 gives pseudo-code for the subroutine Compose called by Solve. The input of Compose is a variable $v \in V$, a state $s$, a partial state $x$, an action set $O$, and a graph $G = (V, E)$ on $V$. Compose generates all sequences from $s$ to states that match $x$ that can be composed using macros for the parents of $v$. We use the Cartesian product $S \times T$ to denote the set of sequences obtained by appending a macro in $T$ to a sequence in $S$. If $s$

---

**Algorithm 4** Compose($v, s, x, O, G$)

1: $S \leftarrow \{\langle\rangle\}$
2: **for all** $w \in Pa(v)$ **do**
3:   **if** $(s \mid V_w) \nsim (x \mid V_w)$ **then**
4:     $T \leftarrow \{\langle s \mid V_w, seq(m), post(m)\rangle \in O : post(m) \sim (x \mid V_w)\}$
5:     **if** $T = \emptyset$ **then**
6:       **return** $\emptyset$
7:     **end if**
8:     $S \leftarrow S \times T$
9:   **end if**
10: **end for**
11: **return** $S$

---

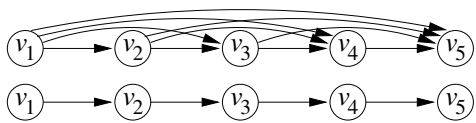

Figure 3: The causal graph and its transitive reduction for $P_5$.

matches $x$, Compose returns a set containing the empty sequence $\langle\rangle$. If, for some parent $w \in Pa(v)$, there is no macro to a state matching $x$, Compose returns the empty set.

### 3.2 Examples

We illustrate MacroPlanner on the well-known Tower of Hanoi problem. An instance of Tower of Hanoi with $n$ discs and 3 pegs can be represented as a planning problem $P_n = \langle V, init, goal, A \rangle$, where $V = \{v_1, \ldots, v_n\}$ are variables representing the discs ($v_1$ being the smallest and $v_n$ the largest), each with domain $D(v_i) = \{A, B, C\}$. The initial state is $init = (v_1 = A, \ldots, v_n = A)$ and the goal state is $goal = (v_1 = C, \ldots, v_n = C)$. For each $v_i \in V$ and each permutation $(j, k, l)$ of $(A, B, C)$, there is an action for moving disc $v_i$ from peg $j$ to peg $k$, formally defined as $a_i^{j,k} = \langle(v_1 = l, \ldots, v_{i-1} = l, v_i = j); (v_i = k)\rangle$. All discs smaller than $v_i$ thus have to be on the third peg $l$ to perform the movement.

Figure 3 shows the causal graph and its transitive reduction for the planning problem $P_5$, representing the 5-disc instance of Tower of Hanoi. Each action for moving disc $v_i$ has a pre-condition on each variable $v_j$, $j < i$. Clearly, the causal graph is acyclic and its transitive reduction is an inverted tree, implying $P_5 \in \mathbf{IR}$. Since the transitive reduction is a directed path, the set of parents of variable $v_i$, $i > 1$, is $Pa(v_i) = \{v_{i-1}\}$. The root variable is $v_5$ or, more generally, $v_n$ for planning problem $P_n$.

To solve Tower of Hanoi, MacroPlanner($P_n$) calls GetMacros($v_n, init, A, G$) to generate the set $M$ of solution macros. In turn, the subroutine GetMacros($v_n, init, A, G$) calls GetMacros($v_{n-1}, init, A, G$). This recursion continues until the base case is reached for GetMacros($v_1, init, A, G$). For each variable $v_i \in V$, let $x_i^d = (v_1 = d, \ldots, v_i = d)$, $d \in \{A, B, C\}$, be the partial state assigning the same value $d$ to each variable in $V_{v_i}$.

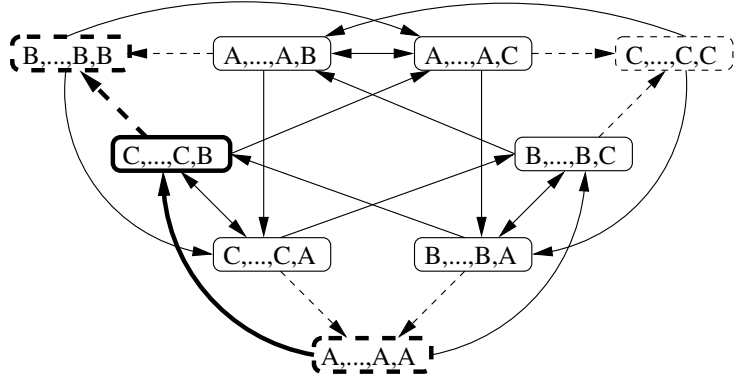

Figure 4: The implicit graph traversed by SOLVE for variable $v_i$.

The set of projected pre-conditions for $v_i$ is $Z = \{x_i^A, x_i^B, x_i^C\}$. Since each projected pre-condition specifies a value for each ancestor of $v_i$, each macro has to end in either of these three partial states. It is easy to show by induction that the set of macros $M$ returned by GETMACROS$(v_i, init, A, G)$ contains precisely 9 macros, corresponding to each pair of projected pre-conditions (including macros with an empty action sequence from a projected pre-condition to itself).

Let us study the behavior of GETMACROS$(v_i, init, A, G)$. Here, we assume that the 9 macros for $v_{i-1}$ in the set $O$ are those described above. The first partial state added to $L$ on line 2 is the projected initial state $x_i^A$. As a result, GETMACROS calls SOLVE$(v_i, x_i^A, Z, O, G)$ on line 10. In the call to SOLVE, the first state-sequence pair added to $Q$ on line 2 is $(x_i^A, \langle \rangle)$. The action $a_i^{A,B}$ has a pre-condition $x_{i-1}^C \oplus (v_i = A)$ that matches $(v_i = x_i^A(v_i)) = (v_i = A)$. As a result, SOLVE calls COMPOSE$(v_i, x_i^A, x_{i-1}^C \oplus (x_i = A), O, G)$ on line 6.

In COMPOSE, there is a single parent of $v_i$, namely $v_{i-1}$. The projection $x_i^A \mid V_{v_{i-1}} = x_{i-1}^A$ is not equal to $x_{i-1}^C$. There is a single macro $m$ in the set $O$ from $x_{i-1}^A$ to $x_{i-1}^C$, so the set $S$ returned by COMPOSE contains a single sequence $\langle m \rangle$. As a result, SOLVE generates a successor state $x_{i-1}^C \oplus (v_i = B)$ of the first type, which is added to $Q$ on line 8 together with its associated sequence $\langle m, a_i^{A,B} \rangle$.

In the iteration for $x_{i-1}^C \oplus (v_i = B)$, there is a single projected pre-condition which matches $(v_i = B)$, namely $x_i^B$, so SOLVE calls COMPOSE$(v_i, x_{i-1}^C \oplus (v_i = B), x_i^B, O, G)$ on line 12. For the only parent $v_{i-1}$, the projection $x_{i-1}^C$ does not match $x_i^B \mid V_{v_{i-1}} = x_{i-1}^B$. There is a single macro $m'$ in the set $O$ from $x_{i-1}^C$ to $x_{i-1}^B$, so the set $S$ returned by COMPOSE contains a single sequence $\langle m' \rangle$. As a result, SOLVE generates a successor state $x_i^B$ of the second type on line 14, and a macro $\langle x_i^A, \langle m, a_i^{A,B}, m' \rangle, x_i^B \rangle$ is added to $M$. The other eight macros are added to $M$ in a similar way. In the next section we prove that MACROPLANNER is guaranteed to generate a solution to Tower of Hanoi in polynomial time.

Figure 4 shows the implicit graph traversed by SOLVE for a variable $v_i$, $i > 1$, in Tower of Hanoi. The path corresponding to the macro from $x_i^A$ to $x_i^B$ in the example is marked in bold. Successor states of the first type, as well as their incoming edges, are denoted by solid lines. Successor states of the second type, as well as their incoming edges, are denoted by dashed lines. Note that the size of the graph is constant and does not depend on $i$. The only difference is the cost of each edge (omitted in the figure), which equals the length of the corresponding action sequence in Solve and depends on the length of the macros for $v_{i-1}$.

Our second example is a planning domain first suggested by Jonsson and Bäckström (1998b). Again, we use $P_n$ to denote the instance containing $n$ variables $v_1, \ldots, v_n$. Each variable $v_i \in V$ has domain $D(v_i) = \{0, 1\}$. The initial state is $(v_1 = 0, \ldots, v_n = 0)$ and the goal state is $(v_1 = 0, \ldots, v_{n-1} = 0, v_n = 1)$. For each variable $v_i$, there are two actions:

$$
\begin{aligned}
a_i &= \langle (v_1 = 0, \ldots, v_{i-2} = 0, v_{i-1} = 1, v_i = 0); (v_i = 1) \rangle, \\
a_i' &= \langle (v_1 = 0, \ldots, v_{i-2} = 0, v_{i-1} = 1, v_i = 1); (v_i = 0) \rangle.
\end{aligned}
$$

The causal graph and its transitive reduction are identical to Tower of Hanoi, so Figure 3 applies here as well, and it follows that $P_n \in \mathbf{IR}$.

Jonsson and Bäckström (1998b) showed that an optimal plan for $P_n$ has length $2^n - 1$. To solve $P_n$ we need to change the value of $v_n$ from 0 to 1 using the action $a_n$. Since the pre-condition of $a_n$ is $(v_{n-1} = 1)$, while the goal state is $(v_{n-1} = 0)$, we need to insert $a_{n-1}$ before $a_n$ and $a_{n-1}'$ after $a_n$, resulting in the sequence $\langle a_{n-1}, a_n, a_{n-1}' \rangle$. Actions $a_{n-1}$ and $a_{n-1}'$ specify the pre-condition $(v_{n-2} = 1)$, while $a_n$ and the goal state specify $(v_{n-2} = 0)$, requiring the value of $v_{n-2}$ to change four times. It is easy to show that the value of variable $v_i$ has to change $2^{n-i}$ times, for a minimum of $2^n - 1$ total actions.

For each variable $v_i \in V$, there are two projected pre-conditions in the set $Z$, namely $(v_1 = 0, \ldots, v_i = 0)$ and $(v_1 = 0, \ldots, v_{i-1} = 0, v_i = 1)$. Since both specify values for each ancestor of $v_i$, the set $M$ of macros returned by GetMacros contains precisely four macros, one for each pair of projected pre-conditions. In spite of the fact that an optimal solution has exponential length in the number of variables, MacroPlanner is guaranteed to generate a solution to $P_n$ in polynomial time due to the complexity results in the next section.

### 3.3 Theoretical Properties

In this section we prove several theoretical properties of the algorithm MacroPlanner. For ease of presentation the proofs of several theorems have been moved to the appendix. The first two theorems are related to the correctness and optimality of MacroPlanner.

**Theorem 3.2.** *For each planning problem $P \in \mathbf{IR}$, each macro $m \in M$ returned by* MacroPlanner$(P)$ *is well-defined and solves $P$.*

The proof of Theorem 3.2 appears in Appendix A. The proof is an induction on variables $v \in V$ to show that each macro returned by Compose and Solve is well-defined. In addition, each macro returned by Solve satisfies a projected pre-condition. For the root variable, Solve is called for the initial state *init*, and the only projected pre-condition is *goal*. The macros returned by MacroPlanner$(P)$ are precisely those returned by this call to Solve, and thus solve $P$.

**Theorem 3.3.** *For each planning problem $P \in \mathbf{IR}$, if there exists a plan solving $P$,* MacroPlanner$(P)$ *returns an optimal plan for $P$, else it returns "fail".*

The proof of Theorem 3.3 appears in Appendix B. The proof is a double induction on variables $v \in V$ and states $s$ for $V_v$ visited during a call to SOLVE, to show that the macros returned by SOLVE represent the shortest solutions to the subproblems corresponding to this call. Thus some macro returned by SOLVE for the root variable has to be an optimal solution for achieving *goal* starting in *init*, which is precisely the definition of an optimal plan for $P$.

We also prove two complexity results for MACROPLANNER. Specifically, we study two subclasses of the class **IR** and prove that plan generation is polynomial for these classes, provided that we are allowed to represent the resulting plan using a hierarchy of macros. For each variable $v \in V$ and each value $d \in D(v)$, we define $A_v^d = \{a \in A : post(a)(v) = d\}$ as the set of actions that change the value of $v$ to $d$ (implying $v \in V_{post(a)}$). We first prove a lemma that relates the complexity of MACROPLANNER to the number of states visited during calls to SOLVE.

**Lemma 3.4.** *Let $g_v$ be the number of states visited during the various calls to* SOLVE *by* GETMACROS$(v, init, A, G)$. *The complexity of* MACROPLANNER *is* $O(|A| \sum_{v \in V} g_v^3)$.

*Proof.* SOLVE is basically a modified version of Dijkstra's algorithm, which is quadratic in the number of nodes of the underlying graph, or $O(g_v^2)$. Each state $p$ is only dequeued from the priority queue $Q$ once. For an action $a \in O$, the call to COMPOSE on line 6 returns a set of distinct action sequences, so $a$ generates at most one edge from $p$ to successor states of the first type. However, two different actions in $O$ may generate two edges from $p$ to the same state. The same is true for projected pre-conditions in $Z$ and successor states of the second type. Since there are at most $|A|$ actions in $O$ and at most $|A|$ projected pre-conditions in $Z$, the worst-case complexity of a single call to SOLVE is $O(|A|g_v^2)$.

The subroutine GETMACROS calls SOLVE for each successor state of the second type. Since there are at most $g_v$ such states, the complexity of GETMACROS is $O(|A|g_v^3)$, so the overall complexity of MACROPLANNER is $O(|A| \sum_{v \in V} g_v^3)$. Note that the complexity of COMPOSE is included in this analysis. COMPOSE is called by SOLVE to generate successor states of either type, so the number $g_v$ of states visited by SOLVE is proportional to the total number of sequences returned by COMPOSE. □

**Theorem 3.5.** *If $V_{pre(a)} = V_v$ for each variable $v \in V \cup \{v^*\}$ and each action $a \in A$ such that $V_{post(a)} = \{v\}$, the complexity of* MACROPLANNER *is* $O(|V||A|^4)$.

**Theorem 3.6.** *Assume that for each variable $v \in V$, each value $d \in D(v)$, and each pair of actions $a, a' \in A_v^d$, $V_{pre(a)} = V_{pre(a')} = \{v\} \cup Pa(v)$ and $pre(a) \mid Pa(v) = pre(a') \mid Pa(v)$. If $pre(a) \mid Pa(v) = init \mid Pa(v)$ for each action $a \in A$ with $post(a)(v) = init(v)$, the algorithm runs in polynomial time with complexity $O(|A| \sum_{v \in V} |D(v)|^3)$.*

The proofs of Theorems 3.5 and 3.6 appear in Appendix C. The proof of Theorem 3.5 is based on the observation that if the pre-conditions are fully specified on the ancestors of a variable $v \in V$, the number of states visited during the various calls to SOLVE is bounded by the number of actions. The proof of Theorem 3.6 follows from the fact that the algorithm always achieves a value $d \in D(v)$ in the domain of $v \in V$ in the same state, so the number of states is linear in the size of the variable domains.

Two examples of planning problems with the properties established in Theorem 3.5 are Tower of Hanoi and the domain suggested by Jonsson and Bäckström (1998b), both from the previous example section.

An example of a planning problem with the properties established in Theorem 3.6 is a domain proposed by Domshlak and Dinitz (2001). The set of variables is $V = \{v_1, \ldots, v_n\}$, each with domain $D(v_i) = \{0, 1, 2\}$, the initial state is $init = (v_1 = 0, \ldots, v_n = 0)$, and the goal state is $goal = (v_1 = 2, \ldots, v_n = 2)$. For each variable $v_i \in V$, there are four actions:

$$
\begin{aligned}
a_i^1 &= \langle (v_{i-1} = 2, v_i = 0); (v_i = 1) \rangle, \\
a_i^2 &= \langle (v_{i-1} = 0, v_i = 1); (v_i = 2) \rangle, \\
a_i^3 &= \langle (v_{i-1} = 2, v_i = 2); (v_i = 1) \rangle, \\
a_i^4 &= \langle (v_{i-1} = 0, v_i = 1); (v_i = 0) \rangle,
\end{aligned}
$$

For $v_1$, the pre-condition on $v_{i-1}$ is dropped. Domshlak and Dinitz (2001) showed that the length of an optimal plan solving the planning problem is $2^{n+1} - 2$. However, due to Theorem 3.6, MacroPlanner generates an optimal solution in polynomial time.

### 3.4 Pruning

To improve the running time, it is possible to prune some states before they are visited. Specifically, let $(p, seq)$ be a state-sequence pair visited during a call to Solve$(v, s, Z, O, G)$. For each state-sequence pair $(t, seq2)$ previously visited during the same call such that $t(v) = p(v)$, attempt to reach $p$ from $t$ using Compose$(v, t, p, O, G)$. Since $p$ is a fully specified state, Compose$(v, t, p, O, G)$ returns at most one action sequence. Call it $seq3$. If $p$ is reachable from $t$ and $\langle seq2, seq3 \rangle$ is at least as short as $seq$, there is no need to visit $p$, since any plan reachable from $s$ via $p$ is also reachable via $t$ on a path of equal or shorter length. The same reasoning holds for macros to a state $p$ that matches a projected pre-condition $z$, under the additional restriction that $t$ also match $z$.

If it is not possible to reach the projected goal state $goal \mid V_v$ from $s$, we can remove all macros whose pre-condition equals $s$ from the set $M$ of macros in GetMacros$(v, init, A, G)$, since no plan that solves the original planning problem $P$ could contain such macros. In addition, we can remove all macros whose post-condition equals $s$ from the set $M$. If it is not possible to reach the projected goal state from the projected initial state $init \mid V_v$, there exists no plan that solves the original planning problem $P$, so we can immediately return "fail" without generating the remaining set of macros for other variables.

## 4. Extending MacroPlanner

In this section, we study ways to extend the algorithm MacroPlanner to a broader class of planning problems. In Subsection 4.1, we propose a relaxed definition of the causal graph that, when applied, extends the class **IR**. We show that MacroPlanner can be modified to solve planning problems in the new class. In Subsection 4.2, we extend MacroPlanner to planning problems with acyclic causal graphs. In Subsection 4.3, we show how to combine the results for different classes to form a more general class of planning problems. Subsection 4.4 illustrates the algorithms for the new classes on two well-known examples: Gripper and Logistics.

## 4.1 Relaxed Causal Graph

The conventional definition of the causal graph $(V, E)$ states that there is an edge $(w, v) \in E$ if and only if $w \neq v$ and there exists an action $a \in A$ such that $w \in V_{pre(a)} \cup V_{post(a)}$ and $v \in V_{post(a)}$. Consider an action $a \in A$ such that $V_{post(a)} = \{w, v\}$. When applied, $a$ changes the values of both $w$ and $v$. Under the conventional definition, $a$ induces a cycle since the causal graph includes the two edges $(w, v)$ and $(v, w)$.

In this section, we propose a relaxed definition of the causal graph. Under the new definition, an acyclic causal graph does not necessarily imply that all actions are unary. Assume that there exists at least one action $a \in A$ such that $V_{post(a)} = \{w, v\}$. Further assume that there are multiple actions for changing the value of $w$, whose pre- and post-conditions do not specify values for $v$. Finally, assume that there exists no action for changing the value of $v$ that does not change the value of $w$.

Under the given assumption, the only way to satisfy the pre-condition of an action that changes the value of $v$ is to first change the value of $w$. This is not changed by the fact that such an action might also change the value of $w$. The opposite is not true of actions that change the value of $w$, since the value of $w$ can change independently of $v$. Under the relaxed definition, the causal graph includes the edge $(w, v)$ but excludes the edge $(v, w)$.

### 4.1.1 DEFINITION OF THE CLASS **RIR**

**Definition 4.1.** *The relaxed causal graph $(V, E')$ of a planning problem $P$ is a directed graph with the variables in $V$ as nodes. There is an edge $(w, v) \in E'$ if and only if $w \neq v$ and either*

1. *there exists $a \in A$ such that $w \in V_{pre(a)} - V_{post(a)}$ and $v \in V_{post(a)}$, or*

2. *there exists $a \in A$ such that $w, v \in V_{post(a)}$ and either*

   (a) *there exists $a' \in A$ such that $w \in V_{post(a')}$ and $v \notin V_{post(a')}$, or*

   (b) *there does not exist $a' \in A$ such that $w \notin V_{post(a')}$ and $v \in V_{post(a')}$.*

Edges of type 2(b) ensure that there always is at least one edge between two variables $w, v \in V$ that appear in the same post-condition of an action $a \in A$. If it is not possible to change the value of either without changing the value of the other, the relaxed causal graph contains a cycle just as before. We can now define the class **RIR** of relaxed inverted tree reducible planning problems:

**Definition 4.2.** *A planning problem $P$ is relaxed inverted tree reducible if and only if its relaxed causal graph is acyclic and the transitive reduction of the relaxed causal graph is an inverted tree.*

Consider a planning problem $P \in$ **IR**. Since each action is unary, the relaxed causal graph contains no edges of type 2. Consequently, the relaxed causal graph is identical to the conventional causal graph, so any tree-reducible planning problem is also relaxed tree-reducible, implying **IR** $\subset$ **RIR**.

As an example, consider a planning problem with just two variables $v$ and $w$. Let $D(v) = \{0, 1, 2, 3\}$, $D(w) = \{0, 1\}$, $init = (v = 0, w = 0)$, and $goal = (v = 3, w = 1)$. Assume that there are three actions:

$$\langle(v=0); (v=1)\rangle,$$
$$\langle(v=2); (v=3)\rangle,$$
$$\langle(v=1, w=0); (v=2, w=1)\rangle.$$

The third action is not unary, since it changes the value of both $v$ and $w$. There are actions for changing the value of $v$ that do not affect $w$, while the opposite is not true. Hence the relaxed causal graph contains a single edge $(v, w)$.

### 4.1.2 Algorithm

We show that a minor modification of MacroPlanner is sufficient to solve planning problems in the class **RIR**. Since actions are no longer unary, an action $a \in A$ that changes the value of a variable $v \in V$ may also change the values of other variables. When generating macros for $v$, we only consider the set $A'$ of actions that change the value of $v$ without changing the value of any descendant of $v$ in the relaxed causal graph. Formally, $A' = \{a \in A : v \in V_{post(a)} \land |V_{post(a)} \cap Desc(v)| = 0\}$.

Just as before, we project the pre-conditions of actions that change the value of some descendant of $v$ onto $V_v$ in order to define subproblems $P_v$ for $v$. Let $A'' = \{a \in A : |V_{post(a)} \cap Desc(v)| > 0 \land (pre(a) \mid V_v) \neq ()\}$ be the set of such actions whose projected pre-condition onto $V_v$ is non-empty. Let $a \in A''$ be such an action, and let $m$ be a macro for $v$ that achieves the pre-condition of $a$, i.e., $post(m) \sim pre(a)$.

The only reason for applying $m$ (unless it also satisfies the projected pre-condition of some other action) is to satisfy the pre-condition of $a$. If $a$ changes the value of $v$, we are no longer in the projected state $post(m)$ as a result of applying $a$. In other words, we should not generate macros for $v$ starting in $post(m)$ as before. Rather, we should generate macros from $post(m) \oplus (post(a) \mid V_v)$, i.e., the projected state that results from applying $a$ in $post(m)$.

---

**Algorithm 5** RelaxedPlanner($P$)

1: $G \leftarrow$ relaxed causal graph of $P$
2: $R \leftarrow$ transitive reduction of $G$
3: $v \leftarrow$ root variable of $R$
4: $M \leftarrow$ RelaxedMacros($v, init, A, R$)
5: **return** $M$, or "fail" if $M = \emptyset$

---

Algorithm 5 shows the modified version of MacroPlanner for the class **RIR**, which we call RelaxedPlanner. The only difference with respect to MacroPlanner is that $G$ now refers to the relaxed causal graph. Algorithm 6 shows the modified version of GetMacros for the class **RIR**, which we call RelaxedMacros. The modifications with respect to GetMacros appear on lines 3–5 and 12–17. The subroutine Solve remains the same as before.

### 4.1.3 Theoretical Properties

**Theorem 4.3.** *For each planning problem $P \in$ **RIR**, each macro $m \in M$ returned by RelaxedPlanner($P$) is well-defined and solves $P$.*

---

**Algorithm 6** RelaxedMacros($v, init, A, G$)

1: $M \leftarrow \emptyset$
2: $L \leftarrow$ list containing the projected initial state $init \mid V_v$
3: $A'' \leftarrow \{a \in A : |V_{post(a)} \cap Desc(v)| > 0 \wedge (pre(a) \mid V_v) \neq ()\}$
4: $Z \leftarrow \{pre(a) \mid V_v : a \in A''\}$
5: $O \leftarrow \{a \in A : v \in V_{post(a)} \wedge |V_{post(a)} \cap Desc(v)| = 0\}$
6: **for all** $u \in Pa(v)$ **do**
7:     $O \leftarrow O \cup$ RelaxedMacros($u, init, A, G$)
8: **end for**
9: **while** there are more elements in $L$ **do**
10:     $s \leftarrow$ next element in $L$
11:     $M \leftarrow M \cup$ Solve($v, s, Z, O, G$)
12:     **for all** $m = \langle pre(m), seq(m), post(m) \rangle \in M$ **do**
13:         **for all** $a \in A'' - \{a^*\}$ such that $post(m) \sim pre(a)$ **do**
14:             **if** $post(m) \oplus (post(a) \mid V_v) \notin L$ **then**
15:                 append $post(m) \oplus (post(a) \mid V_v)$ to $L$
16:             **end if**
17:         **end for**
18:     **end for**
19: **end while**
20: **return** $M$

---

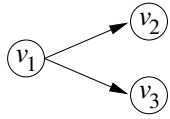

Figure 5: Transitive reduction graph with outdegree $> 1$.

**Theorem 4.4.** *For each planning problem $P \in$ **RIR**, if there exists a plan solving $P$, RelaxedPlanner returns an optimal plan for $P$, else it returns "fail".*

The proofs of Theorems 4.3 and 4.4 are simple adaptations of the corresponding proofs for MacroPlanner, and appear in Appendix D.

## 4.2 Acyclic Causal Graph

In this section we present a second extension of MacroPlanner, this time to the class of planning problems with acyclic causal graphs. In other words, variables in the causal graph may have unbounded outdegree, even when considering the transitive reduction. Unbounded outdegree poses a challenge to the macro compilation approach, as illustrated by the following example.

Consider a planning problem $P$ with $V = \{v_1, v_2, v_3\}$, $D(v_1) = \{0, 1, 2\}$, and $D(v_2) = D(v_3) = \{0, 1\}$. The initial state is $(v_1 = 0, v_2 = 0, v_3 = 0)$, the goal state is $(v_2 = 1, v_3 = 1)$, and $A$ contains the following actions:

$$
\begin{aligned}
a_1^1 &= \langle v_1 = 0; v_1 = 1 \rangle, \\
a_1^2 &= \langle v_1 = 0; v_1 = 2 \rangle, \\
a_2^1 &= \langle v_1 = 1, v_2 = 0; v_2 = 1 \rangle, \\
a_3^1 &= \langle v_1 = 2, v_3 = 0; v_3 = 1 \rangle.
\end{aligned}
$$

Figure 5 shows the causal graph of $P$, identical to its transitive reduction in this case. Since variable $v_1$ has outdegree 2, it follows that $P \notin \mathbf{IR}$ (implying $P \notin \mathbf{RIR}$ since actions are unary). From the point of view of variable $v_2$, it is possible to set $v_2$ to 1: apply $a_1^1$ to change $v_1$ from 0 to 1, followed by $a_2^1$ to change $v_2$ from 0 to 1. From the point of view of $v_3$, it is possible to set $v_3$ to 1 by applying $a_1^2$ followed by $a_3^1$. Nevertheless, there is no valid plan for solving the planning problem.

The algorithm MacroPlanner solves planning problems by decomposing them into subproblems for each variable. We have seen that this is efficient for certain subclasses of $\mathbf{IR}$, and that it guarantees optimality. However, for variables with unbounded outdegree it is no longer possible to define subproblems for each state variable in isolation. If we run MacroPlanner on the example planning problem above, it will generate a macro for setting $v_2$ to 1, and a macro for setting $v_3$ to 1. This might lead us to believe that $P$ has a solution, while in fact it does not.

### 4.2.1 Definition of the Class $\mathbf{AR}$

To extend MacroPlanner to planning problems with acyclic causal graphs, we impose an additional restriction on planning problems. For each state variable $v \in V$ with outdegree greater than 1 in the causal graph, we require $v$ to be reversible.

**Definition 4.5.** *A state variable $v \in V$ is reversible if and only if, for each state $s$ for $V_v$ that is reachable from the projected initial state init $| V_v$, the projected initial state init $| V_v$ is reachable from $s$.*

**Definition 4.6.** *A planning problem $P$ belongs to the class $\mathbf{AR}$ if the causal graph of $P$ is acyclic and each variable is reversible.*

### 4.2.2 Algorithm

In this section we show how to modify the algorithm MacroPlanner so that it solves planning problems in $\mathbf{AR}$. We first describe an algorithm called ReversiblePlanner which requires all variables to be reversible. ReversiblePlanner, which appears in Algorithm 7, operates directly on the causal graph of the planning problem $P$. Thus, it does not compute the transitive reduction of the causal graph. Another feature is that the subroutine Get-Macros is omitted. Instead, ReversiblePlanner directly calls ReversibleCompose, the equivalent of the subroutine Compose.

Algorithm 8 describes the subroutine ReversibleCompose. While Compose returns a set of sequences for reaching a partial state $x$, ReversibleCompose only returns a pair of sequences. Here, $U$ is the set of variables that appear in the post-condition of the action whose pre-condition $x$ we wish to satisfy. The first sequence $seq$ satisfies $x$

---

**Algorithm 7** ReversiblePlanner($P$)

---

1: $G \leftarrow$ [relaxed] causal graph of $P$
2: $(seq, seq2) \leftarrow$ ReversibleCompose($\emptyset, init, goal, A, G$)
3: **return** $seq$

---

**Algorithm 8** ReversibleCompose($U, s, x, A, G$)

---

1: $seq \leftarrow \langle \rangle$, $seq2 \leftarrow \langle \rangle$
2: **for all** $w \in V_x$ in topological order **do**
3:    **if** $x(w) \neq s(w)$ **then**
4:       $m \leftarrow$ ReversibleSolve($w, s, x(w), A, G$)
5:       **if** $m =$ "fail" **then**
6:          **return** ("fail","fail")
7:       **else**
8:          $seq \leftarrow \langle m, seq \rangle$
9:          **if** $w \notin U$ **then**
10:             $m' \leftarrow$ ReversibleSolve($w, s \oplus (w = x(w)), s(w), A, G$)
11:             $seq2 \leftarrow \langle seq2, m' \rangle$
12:          **end if**
13:       **end if**
14:    **end if**
15: **end for**
16: **return** $(seq, seq2)$

---

starting in the state $s$. If this fails, ReversibleCompose returns the pair of sequences ("fail","fail"). Otherwise, the second sequence $seq2$ returns to $s$ from the post-condition of the first sequence when applied in $s$. An exception occurs for variables in $U \subseteq V$, which are not returned to their values in $s$.

ReversibleCompose requires a topological sort of the variables of the causal graph $G$, which can be obtained in polynomial time for acyclic graphs. An example of a topological sort is $v_1, v_2, v_3$ for the causal graph in Figure 5. The values of $x$ are satisfied in reverse topological order because of the way the sequence $seq$ is constructed, i.e., each new macro $m$ is inserted first. ReversibleCompose calls ReversibleSolve to obtain a macro for changing the value of a variable $w \in V_x$ from $s(w)$ to $x(w)$. If this succeeds and $w \notin U$, ReversibleCompose calls ReversibleSolve a second time to obtain a macro for resetting the value of $w$ from $x(w)$ to $s(w)$. This macro is appended to the sequence $seq2$, which resets the values of variables in $V_x - U$ in topological order.

Algorithm 9 describes ReversibleSolve, the equivalent of the subroutine Solve. Unlike Solve, ReversibleSolve only generates a single macro for changing the value of $v$ from $s(v)$ to $d$. Moreover, ReversibleSolve does not invoke Dijkstra's algorithm; instead, it performs a breadth-first search over the values in the joint domain of $W$. A variable $w \in V_v$ belongs to $W$ if there exists an action for changing the value of $v$ or one of its descendants with $w$ in its post-condition. Note that if actions are unary, $W = \{v\}$, so the breadth-first search is over values in the domain $D(v)$ of $v$.

---

**Algorithm 9** ReversibleSolve$(v, s, d, A, G)$

1: $W \leftarrow \{w \in V_v : \exists a \in A$ such that $w \in V_{post(a)}$ and $V_{post(a)} \cap (Desc(v) \cup \{v\}) \neq \emptyset\}$
2: $A' \leftarrow \{a \in A : v \in V_{post(a)} \ [\wedge Desc(v) \cap V_{post(a)} = \emptyset] \}$
3: $L \leftarrow$ list containing the state-sequence pair $(s \mid W, \langle \rangle)$
4: **while** $L$ contains more elements **do**
5: $\quad (p, seq) \leftarrow$ next element in $L$
6: $\quad$**if** $p = (s \mid W) \oplus (v = d)$ **then**
7: $\quad\quad$**return** $\langle (s \mid V_v, seq, (s \mid V_v) \oplus (v = d) \rangle$
8: $\quad$**end if**
9: $\quad$**for all** $a \in A'$ such that $pre(a) \sim (v = p(v))$ **do**
10: $\quad\quad (seq2, seq3) \leftarrow$ ReversibleCompose$(V_{post(a)}, s \oplus p, pre(a), A, G)$
11: $\quad\quad$**if** $seq2 \neq$ "fail" **and** $\not\exists (p', seq') \in L$ such that $p' = p \oplus post(a)(v)$ **then**
12: $\quad\quad\quad$ insert $(p \oplus post(a)(v), \langle seq, seq2, a, seq3 \rangle)$ into $L$
13: $\quad\quad$**end if**
14: $\quad$**end for**
15: **end while**
16: **return** "fail"

---

Each partial state $p$ visited during search is associated with a sequence $seq$ used to arrive at $p$. If $p = (s \mid W) \oplus (v = d)$, ReversibleSolve returns a macro corresponding to the given sequence. Note that this macro only changes the value of $v$, leaving the values of all other variables unchanged. Otherwise, ReversibleSolve tries all actions for changing the value of $v$ whose pre-condition is compatible with $(v = p(v))$. To satisfy the pre-condition of an action $a$ from the state $s \oplus p$, ReversibleSolve calls ReversibleCompose. If ReversibleCompose returns a legal pair of sequences $(seq2, seq3)$ and the partial state $p \oplus post(v)(a)$ has not been previously visited, ReversibleSolve generates a new state-sequence pair $(p \oplus post(v)(a), \langle seq, seq2, a, seq3 \rangle)$. Since $U = V_{post(a)}$, $seq3$ resets all variables to their values in $s \oplus p$ except those whose values are changed by $a$.

Since an equivalent of GetMacros is omitted, ReversibleMacros is goal-driven in the sense that ReversibleSolve is only called whenever needed by the subroutine ReversibleCompose. We use memoization to cache the results of ReversibleCompose and ReversibleSolve to avoid recomputing the same result more than once. The expressions within square brackets indicate the changes that are necessary for the algorithm to work on planning problems with acyclic relaxed causal graphs. Note, however, that ReversiblePlanner is incomplete for planning problems in **AR** with non-unary actions.

### 4.2.3 Theoretical Properties

**Theorem 4.7.** *For each planning problem $P \in$ **AR**, if the sequence seq returned by* ReversiblePlanner$(P)$ *is different from "fail", seq is well-defined for init and satisfies* $(init \oplus post(seq)) \sim goal$.

**Theorem 4.8.** *For each planning problem $P \in$ **AR** with unary actions, if $P$ is solvable,* ReversiblePlanner$(P)$ *returns a sequence seq solving $P$, else it returns "fail".*

**Theorem 4.9.** *For each planning problem $P \in \mathbf{AR}$, the complexity of* ReversiblePlan-
ner *is* $O(\mathcal{D}^k|A|(\mathcal{D}^{k+1} + |V|))$, *where* $\mathcal{D} = max_{v \in V} D(v)$ *and* $k = max_{v \in V}|W|$.

The proofs of Theorems 4.7, 4.8, and 4.9 appear in Appendix E. The proofs of Theorems 4.7 and 4.8 are similar to those for MacroPlanner, and show by induction on variables $v \in V$ that the macros returned by ReversibleCompose and ReversibleSolve are well-defined. However, instead of optimal plans, ReversiblePlanner returns a plan if and only if one exists.

The proof of Theorem 4.9 is based on the fact that in each call to ReversibleSolve, state variables not in $V_v - W$ always take on their initial values. This bounds the total number of states in calls to ReversibleSolve. Note that Theorem 4.9 implies that the complexity of ReversiblePlanner is $O(\mathcal{D}|A|(\mathcal{D}^2 + |V|))$ for planning problems in $\mathbf{AR}$ with unary actions, since $|W| = 1$ for each $v \in V$ in this case. In general, if each of the sets $W$ are fixed, the complexity of ReversiblePlanner is polynomial. Also note that Tower of Hanoi belongs to $\mathbf{AR}$ since each variable is reversible, so planning problems in $\mathbf{AR}$ may have exponentially long solutions.

## 4.3 The Class AOR

We can now combine the two results for the classes $\mathbf{IR}$ and $\mathbf{AR}$. The idea is to use the original algorithm to exhaustively generate macros for each variable $v$, as long as the outdegree of $v$ in the transitive reduction of the causal graph is less than or equal to 1. Whenever we encounter a variable $v$ with outdegree larger than 1, we switch to the algorithm for reversible variables from the previous section. This way, we can handle acyclic causal graphs even when some variables are not reversible. We call the resulting class $\mathbf{AOR}$.

### 4.3.1 Definition of the Class AOR

**Definition 4.10.** *A planning problem $P$ belongs to the class $\mathbf{AOR}$ if the causal graph of $P$ is acyclic and each variable $v \in V$ with outdegree > 1 in the transitive reduction of the causal graph is reversible.*

The example planning problem from the previous section does not belong to the class $\mathbf{AOR}$, since variable $v_1$ with outdegree 2 is not reversible. However, assume that we add the following two actions to $A$:

$$a_1^3 = \langle v_1 = 1; v_1 = 0 \rangle,$$
$$a_1^4 = \langle v_1 = 2; v_1 = 0 \rangle.$$

With the new actions, variable $v_1$ is reversible since each value of $v_1$ is reachable from each other value. Since $v_1$ is the only state variable with outdegree greater than 1 in the transitive reduction, $P \in \mathbf{AOR}$.

### 4.3.2 Algorithm

In this section, we describe AcyclicPlanner, which combines ideas from MacroPlanner and ReversiblePlanner to solve planning problems in the class $\mathbf{AOR}$. Acyclic-Planner appears in Algorithm 10. Just like MacroPlanner, AcyclicPlanner operates

---

**Algorithm 10** AcyclicPlanner($P$)

---

1: $G \leftarrow$ [relaxed] causal graph of $P$
2: $R \leftarrow$ transitive reduction of $G$
3: $(S, seq, seq2) \leftarrow$ AcyclicCompose($v^*, init, goal, A, G, R$)
4: **if** $S = \emptyset$ **then**
5:     **return** "fail"
6: **else**
7:     **return** $S \times seq$, or $\langle seq', seq \rangle$ for some $seq' \in S$
8: **end if**

---

on the transitive reduction $R$ of the causal graph. However, since ReversiblePlanner operates on the causal graph $G$ itself, AcyclicPlanner passes on both $G$ and $R$ to the subroutine AcyclicCompose.

---

**Algorithm 11** AcyclicCompose($v, s, x, A, G, R$)

---

1: $S \leftarrow \{\langle \rangle\}$
2: $x' \leftarrow x$ projected onto non-reversible variables
3: **for all** $w \in Pa(v)$ **do**
4:     **if** $outdegree(R, w) \leq 1$ **and** $(s \mid V_w) \not\sim (x' \mid V_w)$ **then**
5:         $M \leftarrow$ AcyclicSolve($w, s \mid V_w, A, G, R$)
6:         $T \leftarrow \{\langle s \mid V_w, seq(m), post(m) \rangle \in M : post(m) \sim (x' \mid V_w)\}$
7:         **if** $T = \emptyset$ **then**
8:             **return** $(\emptyset, "fail", "fail")$
9:         **end if**
10:         $S \leftarrow S \times T$
11:     **end if**
12: **end for**
13: $seq \leftarrow \langle \rangle$, $seq2 \leftarrow \langle \rangle$
14: **for all** $w \in V_x$ in reverse topological order **do**
15:     **if** $outdegree(R, w) > 1$ **and** $(s \mid V_w) \not\sim (x \mid V_w)$ **then**
16:         $(seq3, seq4) \leftarrow$ ReversibleCompose($\emptyset, s \mid V_w, x, A, G$).
17:         **if** $seq3 = $ "fail" **then**
18:             **return** $(\emptyset, "fail", "fail")$
19:         **else**
20:             $seq \leftarrow \langle seq, seq3 \rangle$
21:             $seq2 \leftarrow \langle seq4, seq2 \rangle$
22:         **end if**
23:     **end if**
24: **end for**
25: **return** $(S, seq, seq2)$

---

As can be seen, AcyclicPlanner calls AcyclicCompose with the dummy variable $v^*$. Note that, contrary to before, the transitive reduction of the causal graph may contain several edges to $v^*$. The reason is that there is not a single root variable of the causal graph.

Instead, there may be multiple sink variables in the causal graph, i.e., variables with no outgoing edges. The transitive reduction contains an edge from each sink variable in $V_{goal}$ to $v^*$.

Algorithm 11 describes the subroutine AcyclicCompose. Although it looks somewhat complex, lines 3 through 12 are really just an adaptation of Compose to the class **AOR**. The only difference is that AcyclicSolve does not generate macros for satisfying the projected pre-condition $x$; instead, it generates macros for satisfying the projection $x'$ of $x$ onto non-reversible variables. The idea is for reversible variables to always remain in their initial values. This way, we can treat the parents $w \in Pa(v)$ of $v$ as if they were independent, since any common ancestors have to have outdegree larger than 1 in the transitive reduction, and are thus reversible by definition.

Lines 13 through 24 have the effect of satisfying the values of $x$ for variables in $V_x$ whose outdegree in the transtitive reduction is larger than 1. This is done simply by calling the subroutine ReversibleCompose from the previous section. AcyclicCompose returns three values: a set $S$ of sequences for satisfying the partial state $x$ with respect to non-reversible variables, a sequence $seq$ for satisfying the partial state $x$ with respect to reversible variables, and a sequence $seq2$ for resetting the reversible variables to their initial values.

Note that once ReversibleCompose is called for a variable $w$, each predecessor of $w$ is processed using ReversibleSolve and ReversibleCompose as well. For this to work, the predecessors of $w$ have to be reversible, which is guaranteed by the following lemma:

**Lemma 4.11.** *If $v$ is reversible, so is every predecessor of $v$.*

*Proof.* By contrapositive. Assume that $w$ is a non-reversible predecessor of $v$. Then there exists a state $s$ for $V_w$ that is reachable from the projected initial state $init \mid V_w$ such that $init \mid V_w$ is not reachable from $s$. Since $w$ is a predecessor of $v$, $V_w$ is a subset of $V_v$, so the state $(init \mid V_v) \oplus s$ is reachable from $init \mid V_v$, but not vice versa. Thus $v$ is not reversible. □

Algorithm 12 describes the subroutine AcyclicSolve. The only difference between Solve and AcyclicSolve is that AcyclicSolve calls AcyclicCompose to obtain sequences for satisfying a projected pre-condition. Moreover, whenever satisfying the pre-condition $pre(a)$ of an action, AcyclicCompose applies the sequence $seq3$ to reset reversible variables to their initial values. In addition, AcyclicCompose does not satisfy a projected pre-condition $z$ for reversible variables, since it does not append the sequence $seq2$ to the resulting sequence. Just as before, the expressions within square brackets are those that need to be added to deal with acyclic relaxed causal graphs.

### 4.3.3 Theoretical Properties

**Theorem 4.12.** *For each planning problem $P \in$ **AOR** with unary actions, if there exists a plan solving $P$, each sequence $seq$ in the set $S$ returned by AcyclicPlanner is well-defined for $init$ and satisfies $(init \oplus post(seq)) \sim goal$.*

*Proof.* To prove Theorem 4.12, it is sufficient to combine results for MacroPlanner and ReversiblePlanner. The subroutines AcyclicCompose and AcyclicSolve are identical to Compose and Solve for variables with outdegree less than or equal to 1, so Lemmas

---

**Algorithm 12** AcyclicSolve($v, s, A, G, R$)

---

1:   $M \leftarrow \emptyset$
2:   $A' \leftarrow \{a \in A : v \in V_{post(a)} \; [\wedge Desc(v) \cap V_{post(a)} = \emptyset] \}$
3:   $Z \leftarrow \{pre(a) \mid V_v : a \in A \wedge |V_{post(a)} \cap Desc(v)| > 0\} - \{()\}$
4:   $Q \leftarrow$ priority queue containing the state-sequence pair $(s, \langle \rangle)$
5:   **while** $Q$ is non-empty **do**
6:     $(p, seq) \leftarrow$ remove highest priority state-sequence pair from $Q$
7:     **for all** $a \in A'$ such that $pre(a) \sim (v = p(v))$ **do**
8:       $(S, seq2, seq3) \leftarrow$ AcyclicCompose($v, p, pre(a), A, G, R$)
9:       **for all** sequences $seq4 \in S$ **do**
10:         insert $(p \oplus post(\langle seq4, seq2, a, seq3 \rangle), \langle seq, seq4, seq2, a, seq3 \rangle)$ into $Q$
11:       **end for**
12:     **end for**
13:     **for all** $z \in Z$ such that $z \sim (v = p(v))$ **do**
14:       $(S, seq2, seq3) \leftarrow$ AcyclicCompose($v, p, z, P, G, R$)
15:       **for all** sequences $seq4 \in S$ **do**
16:         $M \leftarrow M \cup \{\langle s, \langle seq, seq4 \rangle, p \oplus post(seq4) \rangle\}$
17:       **end for**
18:     **end for**
19:   **end while**
20:   **return** $M$

---

A.1 and A.2 imply that the resulting sequences and macros have the desired properties. For variables with outdegree larger than 1, ReversibleCompose returns sequences with the desired properties due to Lemma E.1. □

**Theorem 4.13.** *Assume that there exists a constant $k$ such that for each non-reversible variable $v \in V$, the number of non-reversible ancestors of $v$ is less than or equal to $k$. In addition, for each reversible variable $v \in V$, $|W| \leq k$, where $W$ is the set defined in* ReversibleSolve. *Then* AcyclicPlanner *runs in polynomial time.*

*Proof.* Again, we can combine results for MacroPlanner and ReversiblePlanner. First note that any call to ReversibleCompose is polynomial due to Theorem 4.9. Lemma 3.4 states that the complexity of MacroPlanner is $O(|A| \sum_{v \in V} g_v^3)$, where $g_v$ is the number of states visited during calls to Solve for $v$. Since AcyclicCompose and Acyclic-Solve are identical to Compose and Solve for non-reversible variables, the argument in the proof of Lemma 3.4 holds for AcyclicSolve as well. The fact that $v$ has at most $k$ non-reversible predecessors implies $g_v = O(\mathcal{D}^{k+1})$, since reversible predecessors always take on their initial values.

Note that the argument in Lemma 3.4 does not take into account the external call to AcyclicCompose by AcyclicPlanner, which may be exponential in the number $|Pa(v^*)|$ of parents of $v^*$. A slight modification of the algorithm is necessary to prove the theorem. We modify the external call so that it only returns a single sequence. In other words, for each $w \in Pa(v^*)$, we only keep one of the macros returned by the call to AcyclicSolve for $w$. Note that this does not alter the admissibility of the solution. □

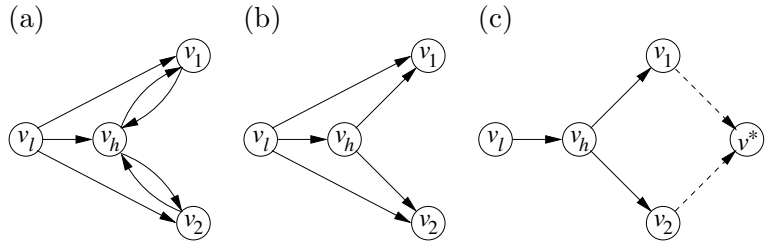

Figure 6: (a) Conventional and (b) relaxed causal graph, and (c) transitive reduction of the relaxed causal graph for GRIPPER.

## 4.4 Examples

In this section, we provide examples of planning problems in **AR** and **AOR**, and describe how REVERSIBLEPLANNER and ACYCLICPLANNER solve them.

### 4.4.1 GRIPPER

Our first example is the well-known GRIPPER domain, in which a robot has to transport balls between two rooms. In GRIPPER, a planning problem $P_n$ is defined by the number $n$ of balls that the robot has to transport. The set of variables is $V = \{v_l, v_h, v_1, \ldots, v_n\}$ with domains $D(v_l) = \{0, 1\}$, $D(v_h) = \{0, 1, 2\}$, and $D(v_i) = \{0, 1, R\}$ for each $1 \leq i \leq n$. The initial state is $init = (v_l = 0, v_h = 0, v_1 = 0 \ldots, v_n = 0)$ and the goal state is $goal = (v_1 = 1, \ldots, v_n = 1)$.

The two rooms are denoted 0 and 1. Variable $v_l$ represents the location of the robot, i.e., either of the two rooms. Variable $v_h$ represents the number of balls currently held by the robot, for a maximum of 2. Finally, variable $v_i$, $1 \leq i \leq n$, represents the location of ball $i$, i.e., either of the two rooms or held by the robot $(R)$. The set $A$ contains the following actions:

$$\langle (v_l = a); (v_l = 1 - a) \rangle, \qquad a \in \{0, 1\},$$
$$\langle (v_l = a, v_h = b, v_i = a); (v_h = b + 1, v_i = R) \rangle, \qquad a, b \in \{0, 1\}, 1 \leq i \leq n,$$
$$\langle (v_l = a, v_h = b + 1, v_i = R); (v_h = b, v_i = a) \rangle, \qquad a, b \in \{0, 1\}, 1 \leq i \leq n.$$

Actions of the first type move the robot between rooms. Actions of the second type cause the robot to pick up a ball at its current location, incrementing the number of balls held. Actions of the third type cause the robot to drop a ball at its current location, decrementing the number of balls held.

For each $1 \leq i \leq n$, there exist actions for changing the value of $v_i$ that also change the value of $v_h$. Consequently, the conventional causal graph contains the edges $(v_h, v_i)$ and $(v_i, v_h)$, inducing cycles. However, there are no actions for changing the value of $v_i$ that do not also change the value of $v_h$. The opposite is not true; there are actions for $v_h$ that do not affect the value of $v_i$ (assuming $n > 1$). Thus, the relaxed causal graph contains the edge $(v_h, v_i)$, but not $(v_i, v_h)$. Figure 6 shows the conventional and relaxed causal graphs, as well as the transitive reduction of the relaxed causal graph, for GRIPPER with $n = 2$. The transitive reduction includes the dummy variable $v^*$ as well as its incoming edges.

The relaxed causal graph is acyclic, and each variable is reversible since we can return to the initial state from any other state. Thus GRIPPER belongs to the class **AR** when considering the relaxed causal graph. Since actions in GRIPPER are non-unary, REVERSIBLEPLANNER is not guaranteed to find a solution. However, we show that it does in fact find a solution to GRIPPER.

REVERSIBLEPLANNER calls REVERSIBLECOMPOSE with the initial and goal states. In turn, REVERSIBLECOMPOSE goes through each variable $w \in V_{goal}$ and calls REVERSIBLE-SOLVE to obtain a macro for setting the value of $w$ to its goal value. Since there are no directed paths between any variables representing balls, a topological sort orders these variables arbitrarily. Without loss of generality, assume that the topological order is $v_1, \ldots, v_n$. Thus the first variable processed by REVERSIBLECOMPOSE is $v_1$. REVERSIBLECOMPOSE calls REVERSIBLESOLVE with $v = v_1$, $s = init$ and $d = 1$.

For $v_1$, the set $W = \{v_h, v_1\}$ contains variables $v_h$ and $v_1$. The set $A'$ contains all actions of the second and third type for picking up and dropping ball 1. The list $L$ initially contains the partial state $(v_h = 0, v_1 = 0)$. There are two actions whose pre-conditions match $(v_1 = 0)$: one for picking up ball 1 from room 0 assuming no ball is currently held $(v_h = 0)$, and one for picking up ball 1 assuming one ball is currently held $(v_h = 1)$. For each of these, REVERSIBLESOLVE calls REVERSIBLECOMPOSE with $U = V_{post(a)}$, $s = init$, and $x = pre(a)$ to obtain a sequence for satisfying the pre-condition of the action.

The pre-condition of the first action is already satisfied in $init$, so REVERSIBLECOMPOSE returns a pair of empty sequences. As a result, REVERSIBLESOLVE adds the partial state $(v_h = 1, v_1 = R)$ to $L$, corresponding to the post-condition of this action. Note that the set $A'$ is empty for variable $v_h$, since there are no actions for changing the value of $v_h$ that do not also change a value of its descendants. Thus it is not possible to satisfy the pre-condition $(v_h = 1)$ of the second action starting in the initial state without changing the value of $v_1$. Consequently, REVERSIBLECOMPOSE returns ("fail","fail"), and no new partial state is added to $L$.

From $(v_h = 1, v_1 = R)$, there are four actions whose pre-conditions match $(v_1 = R)$: two for dropping ball 1 in room 0, and two for dropping ball 1 in room 1. Two of these have pre-condition $(v_h = 2)$ which is impossible to satisfy without changing the value of $v_1$. Dropping ball 1 in room 0 returns to the initial state, so no new partial states are added to $L$. Dropping ball 1 in room 1 requires a pre-condition $(v_l = 1, v_h = 1)$. Since $(v_h = 1)$ already holds, REVERSIBLECOMPOSE calls REVERSIBLESOLVE to change the value of $v_l$ from 0 to 1. This is possible using the action for changing the location of the robot. REVERSIBLECOMPOSE also returns a sequence for returning the robot to room 0.

As a result, REVERSIBLESOLVE adds the partial state $(v_h = 0, v_1 = 1)$ to $L$, corresponding to the post-condition of this action. Note that this partial state satisfies $(v_h = 0, v_1 = 1) = (v_h = 0, v_1 = 0) \oplus (v_1 = 1)$, so REVERSIBLESOLVE returns a macro that changes the value of $v_1$ without changing the values of any other variables. Since REVERSIBLESOLVE successfully returned a macro, REVERSIBLECOMPOSE calls REVERSIBLE-SOLVE again to obtain a macro for returning ball 1 to its initial location. This continues for the other balls. The final solution contains macros for changing the location of each ball from room 0 to room 1 one at a time.

Note that the complexity of REVERSIBLESOLVE is polynomial for GRIPPER due to Theorem 4.9, since $|W| \leq 2$ for each $v_i$, $1 \leq i \leq n$. Also note that the solution is not optimal, since the robot could carry two balls at once. We could also solve GRIPPER using ACYCLICPLANNER, which would treat variables $v_i$, $1 \le i \le n$, as non-reversible since they have outdegree 0 in the transitive reduction of the relaxed causal graph. The solution is in fact the same in this case, since there is just a single way to move a ball from room 0 to room 1.

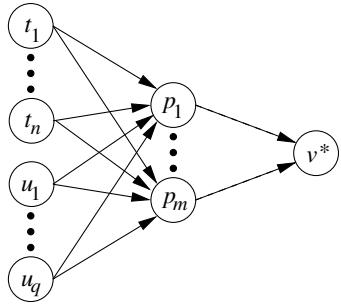

Figure 7: Causal graph and transitive reduction for LOGISTICS.

### 4.4.2 LOGISTICS

Our second example is the LOGISTICS domain, in which a number of packages have to be moved to their final location using trucks and airplanes. Let $m$ denote the number of packages, $n$ the number of trucks, and $q$ the number of airplanes. A multi-valued variable formulation of a planning problem in LOGISTICS is given by $V = \{p_1, \ldots, p_m\} \cup T$, where $T = \{t_1, \ldots, t_n, u_1, \ldots, u_q\}$, $p_i$, $1 \le i \le m$, are packages, $t_j$, $1 \le j \le n$, are trucks, and $u_k$, $1 \le k \le q$, are airplanes.

Let $L$ be a set of possible locations of packages. There is a partition $L_1 \cup \ldots \cup L_r$ of $L$ such that each $L_l$, $1 \le l \le r$, corresponds to locations in the same city. There is also a subset $C$ of $L$ corresponding to airports, with $|C \cap L_l| \ge 1$ for each $1 \le l \le r$. For each $p_i$, the domain $D(p_i) = L \cup T$ corresponds to all possible locations plus inside all trucks and airplanes. For each $t_j$, $D(t_j) = L_l$ for some $1 \le l \le r$. For each $u_k$, $D(u_k) = C$. The initial state is an assignment of values to all variables, and the goal state is an assignment to the packages $p_i$ among the values $L$. The set $A$ contains the following three types of actions:

$$\langle (t = l_1); (t = l_2) \rangle, \quad t \in T, \, l_1 \ne l_2 \in D(t),$$
$$\langle (t = l, p_i = l); (p_i = t) \rangle, \quad t \in T, \, l \in D(t), \, 1 \le i \le m,$$
$$\langle (t = l, p_i = t); (p_i = l) \rangle, \quad t \in T, \, l \in D(t), \, 1 \le i \le m.$$

The first type of action allows a truck or airplane to move between two locations in its domain. The second type of action loads a package in a truck or airplane at the same location. The third type of action unloads a package from a truck or airplane.

Figure 7 shows the causal graph and its transitive reduction for a planning problem in LOGISTICS. The transitive reduction is identical to the causal graph in this case. Just as for GRIPPER, the figure shows the dummy variable $v^*$ as well as its incoming edges. It is easy to see that the causal graph is acyclic. In addition, all variables are reversible since it is possible to return to the initial state from any other state. Thus we could apply REVERSIBLEPLANNER to solve planning problems in LOGISTICS.

| Discs | Time (ms) | Macros | Length |
|-------|-----------|--------|--------|
| 10 | 31 | 27/82 | $> 10^3$ |
| 20 | 84 | 57/172 | $> 10^6$ |
| 30 | 161 | 87/262 | $> 10^9$ |
| 40 | 279 | 117/352 | $> 10^{12}$ |
| 50 | 457 | 147/442 | $> 10^{15}$ |
| 60 | 701 | 177/532 | $> 10^{18}$ |

Table 1: Results in Tower of Hanoi.

To solve a planning problem in LOGISTICS, REVERSIBLEPLANNER calls REVERSIBLE-COMPOSE with the initial and goal states. Since there are no directed paths between any variables representing the location of packages, a topological sort orders these variables arbitrarily. Without loss of generality, assume that the topological order is $p_1, \ldots, p_n$. Thus the first variable processed by REVERSIBLECOMPOSE is $p_1$. REVERSIBLECOMPOSE calls REVERSIBLESOLVE with $v = p_1$, $s = init$ and $d$ equal to the goal location of the package.

For $p_1$, the set $W = \{p_1\}$ just contains $p_1$ since actions are unary. The set $A'$ contains any actions for picking up or dropping the package. Although there may be many ways to move a package from its initial location to its goal location, REVERSIBLESOLVE just considers once such way of moving the package. After moving the package, all trucks and airplanes are returned to their initial locations. REVERSIBLECOMPOSE also calls REVERSIBLESOLVE to generate a macro for returning the package to its initial location. The final solution consists of macros for moving each package to its final location one at a time.

Note that the complexity of REVERSIBLESOLVE is polynomial for LOGISTICS due to Theorem 4.9, since $|W| \leq 1$ for each $p_i$, $1 \leq i \leq m$. Also note that the solution is not optimal, since trucks and airplanes are always returned to their initial locations. We could also solve LOGISTICS using ACYCLICPLANNER, which would treat variables $p_i$, $1 \leq i \leq m$, as non-reversible since they have outdegree 0 in the transitive reduction of the relaxed causal graph. Again, the solution is the same since each ancestor of $p_i$ is reversible, so the states in ACYCLICSOLVE always have trucks and airplanes in their initial locations.

## 5. Experimental Results

To test our algorithm, we ran experiments in two domains: Tower of Hanoi and an extended version of GRIPPER. The results of the experiments in Tower of Hanoi appear in Table 1. We varied the number of discs in increments of 10 and recorded the running time of MACROPLANNER. The table also shows the number of macros generated by our algorithm, and out of those, how many were used to represent the resulting global plan. For example, 27 out of 82 generated macros formed part of the solution to Tower of Hanoi with 10 discs.

For the second set of experiments, we modified the version of GRIPPER from the previous section. Instead of two rooms, the environment consists of a maze with 967 rooms. To transport balls, the robot must navigate through the maze from the initial location to the goal location. The robot can only pick up and drop balls at the initial and goal locations. The results of running ReversiblePlanner on the corresponding planning problems appear in Table 2. Since ReversiblePlanner only generates macros as necessary, all of them are used in the solution. The experiments illustrate the benefit of using macros to store the solution for navigating through the maze. Any algorithm that does not use macros has to recompute a path through the maze every time it goes to pick up a new ball.

| Balls | Time (ms) | Macros | Length |
|-------|-----------|--------|--------|
| $10^0$ | 1528 | 3 | $\approx 300 \times 10^0$ |
| $10^1$ | 1535 | 14 | $\approx 300 \times 10^1$ |
| $10^2$ | 1570 | 104 | $\approx 300 \times 10^2$ |
| $10^3$ | 2906 | 1004 | $\approx 300 \times 10^3$ |

Table 2: Results in Gripper.

## 6. Related Work

For ease of presentation, we group related work into three broad categories: complexity results, macro-based planning, and factored planning. Each of the following subsections presents related work in one of these categories.

### 6.1 Complexity Results

Early complexity results for planning focused on establishing the hardness of different formulations of the general planning problem. If the STRIPS formalism is used, deciding whether a solution exists is undecidable in the first-order case (Chapman, 1987) and PSPACE-complete in the propositional case (Bylander, 1994). For PDDL, the representation language used at the International Planning Competition, the fragment including delete lists is EXPSPACE-complete (Erol, Nau, & Subrahmanian, 1995).

Recently, several researchers have studied tractable subclasses of planning problems that can be provably solved in polynomial time. Most of these subclasses are based on the notion of a causal graph (Knoblock, 1994), which models the degree of independency between the problem variables. However, Chen and Giménez (2008a) showed that any connected causal graph causes the problem to be hard (unless established assumptions such as P = NP fail), so additional restrictions on the problem are necessary.

A common restriction is that the variables of the problem are binary. Jonsson and Bäckström (1998a) defined the class 3S of planning problems with acyclic causal graphs and binary variables. In addition, variables are either static, symmetrically reversible, or splitting. The authors showed that it is possible to determine in polynomial time whether or not a solution exists, although solution plans may be exponentially long.

Giménez and Jonsson (2008) designed a macro-based algorithm that solve planning problems in 3S in polynomial time. The algorithm works by generating macros that change the value of a single variable at a time, maintaining the initial values for the ancestors of the variable. In fact, this is the same strategy used in our algorithm ReversiblePlanner, which can thus be seen as an extension of their algorithm to multi-valued reversible variables and general acyclic causal graphs.

Brafman and Domshlak (2003) designed a polynomial-time algorithm for solving planning problems with binary variables and polytree acyclic causal graphs of bounded indegree. Giménez and Jonsson (2008) showed that if the causal graph has unbounded indegree, the problem becomes NP-complete. Restricting the pre-condition of each action to be on at most two variables causes the problem class to become tractable again (Katz & Domshlak, 2008a). Since the hardness proof is based on a reduction to planning problems with inverted tree causal graphs, these results all apply to the class **IR** if restricting to inverted tree reducible planning problems.

Another result that applies to the class **IR** is the hardness of planning problems with directed path causal graphs (Giménez & Jonsson, 2008). Giménez and Jonsson (2009) extended this result to multi-valued variables with domains of size at most 5. Katz and Domshlak (2008b) showed that planning problems whose causal graphs are inverted forks are tractable as long as the root variable has a domain of fixed size. This class is also a fragment of **IR**. In other words, the class **IR** includes several known tractable and intractable fragments, although the tractability results proven in the present paper are novel.

Other tractability results include the work by Haslum (2008), who defined planning problems in terms of graph grammars, and showed that the resulting class is tractable under certain restrictions of the grammar. Chen and Giménez (2007) defined the width of planning problems and designed an algorithm for solving planning problems whose complexity is exponential in the width. In other words, planning problems with constant width are tractable.

The idea of using reversible variables is related to the work by Williams and Nayak (1997), who designed a polynomial-time algorithm for solving planning problems with acyclic causal graphs and reversible actions. In other words, each action has a symmetric counterpart with the same pre-condition except that it reverses the effect of the former. In addition, their algorithm requires that no two actions have pre-conditions such that one is a proper subset of the other. Our definition of reversible variables is more flexible and does not require actions to be reversible.

Since the class **IR** and the associated algorithm were introduced, two other tractability results based on macros have appeared in the literature. We have already mentioned the work by Giménez and Jonsson (2008) for the class 3S. Also, Chen and Giménez (2008b) presented a polynomial-time algorithm that generates all macros within Hamming distance $k$ of a state, and defined an associated tractable class of planning problems with constant Hamming width. Their class also contains Tower of Hanoi, but its relationship to **IR** has not been fully established.

## 6.2 Macro-Based Planning

It is also worth mentioning the relationship to other macro-based algorithms for planning. Macros have been long used in planning, beginning with the advent of the STRIPS representation (Fikes & Nilsson, 1971). Minton (1985) and Korf (1987) developed the idea further, the latter showing that macros can exponentially reduce the search space if chosen carefully. (Knoblock, 1994) developed an abstraction technique similar to that of macros, where the problem is treated at different levels of abstractions until fully solved. Vidal (2004) extracted macros from relaxed plans used to generate heuristics.

Methods such as Macro-FF (Botea, Enzenberger, Müller, & Schaeffer, 2005) automatically generate macros that are experimentally shown to be useful for search, which proved competitive at the fourth International Planning Competition. Typically, the macros introduced are flat sequences of two or three actions. This stands in stark contrast to our algorithms, in which macros may be hierarchical and have arbitrarily long sequences. In other words, the two approaches are very different: one tries to augment PDDL with slightly longer sequences in the hope of speeding up search, while the other attempts to generate some or all macros that are needed to solve the subproblems associated with a variable. Recent work on macros admits longer action sequences (Botea, Muller, & Schaeffer, 2007; Newton, Levine, Fox, & Long, 2007).

## 6.3 Factored Planning

Another related field of work is factored planning, which attempts to decompose a planning problem into one or several subproblems. Typically, a planning problem is factored into several subdomains, which are organized in a tree structure. Each variable and action of the problem belongs to one of the subdomains. Amir and Engelhardt (2003) introduced an algorithm called PartPlan that solves planning problems for which a tree decomposition already exists. The algorithm is exponential in the maximum number of actions and variables of a subdomain.

Brafman and Domshlak (2006) introduced an algorithm called LID-GF which decomposes planning problems based on the causal graph. LID-GF is polynomial for planning problems with fixed local depth and causal graphs of fixed tree-width. The local depth of a variable is defined as the number of times that the value of the variable has to change on a plan solving the problem. Interestingly, Tower of Hanoi, which is solved in polynomial time by MacroPlanner, has exponential local depth and a causal graph with unbounded tree-width. Kelareva, Buffet, Huang, and Thiébaux (2007) proposed an algorithm for factored planning that automatically chooses the order of solving subproblems. The algorithm requires a subproblem clustering to be given, while our algorithms automatically derives a subproblem order.

## 7. Conclusion

In this paper, we have introduced the class **IR** of inverted tree reducible planning problems and presented an algorithm called MacroPlanner that uses macros to solve planning problems in this class. The algorithm is provably complete and optimal, and runs in polynomial time for several subclasses of **IR**.

We also extended the class **IR** in several ways. The class **RIR** allows the causal graph to be relaxed, and an associated algorithm called RelaxedPlanner is able to solve planning problems in **RIR** optimally. The class **AR** allows general acyclic causal graphs as long as variables are reversible, and an associated algorithm called ReversiblePlanner can solve planning problems with unary actions in **AR** non-optimally in polynomial time. For non-unary actions, ReversiblePlanner is not complete, although it can still solve some problems such as GRIPPER. Finally, the class **AOR** allows acyclic causal graphs as long as each variable with outdegree larger than 1 in the transitive reduction is reversible. The algorithm ACYCLICPLANNER combines ideas from the other algorithms to solve planning problems in the class **AOR**.

We validated the theoretical properties of the algorithm in two sets of experiments. In the first set, the algorithm was applied to Tower of Hanoi. Although the optimal solution is exponential, the algorithm is guaranteed to solve problems in the domain in polynomial time. The reason is that macros obviate the need to solve the same subproblem multiple times. In the second set of experiments, we applied the algorithm to an extended version of GRIPPER. Again, the results demonstrate the power of using macros to store the solution to subproblems.

A major contribution of the paper is extending the set of known tractable classes of planning problems. Katz and Domshlak (2008b) suggested projecting planning problems onto known tractable fragments in order to compute heuristics. This is perfectly possible to do using our algorithms. In the classes **IR** and **RIR**, the resulting heuristic will be admissible since the corresponding solution is optimal. Even if the solution is exponentially long, computing the solution length can be done in polynomial time using dynamic programming.

If two domains share part of the causal graph structure, macros generated in one domain could be used in the other. This could save significant computational effort since the scheme compiles away variables and replaces them with macros. Since macros are functionally equivalent to standard actions, they can be used in place of the actions and variables they replace. Macros can also prove useful in domains for which we have to backtrack to find an optimal solution. If macros have already been generated and stored for some variables, there is no need to recompute partial plans for these variables from scratch.

If the multi-valued representation included a notion of objects, it would be possible to generate macros for one object and share those macros among identical objects, just like the parametrized operators of a standard PDDL planning domain. For example, in the LOGISTICS domain, each truck that operates in the same city is functionally equivalent, and a macro generated for one truck can be applied to another. The same is true for airplanes. Finally, to obtain optimal plans for the classes **AR** and **AOR**, one would have to test all ways to interleave subplans for different variables, a process which is not likely to be tractable unless the number of interleaved subplans is constant.

## Acknowledgments

This work was partially funded by APIDIS and MEC grant TIN2006-15387-C03-03.

## Appendix A. Proof of Theorem 3.2

In this appendix we prove Theorem 3.2, which states that each macro $m$ returned by MACROPLANNER($P$) is well-defined and solves $P$. First, we prove a series of lemmas which state that the sequences and macros returned by COMPOSE and SOLVE are well-defined.

**Lemma A.1.** *For each $v \in V$, each state $s$ for $V_v$, and each partial state $x$ for $V_v$ such that $x \sim (v = s(v))$, if each macro for the parents of $v$ is well-defined, each action sequence seq returned by* COMPOSE($v, s, x, O, G$) *is well-defined for $s$ and $(s \oplus post(seq)) \sim x$.*

*Proof.* Compose$(v, s, x, O, G)$ returns sequences of the type $seq = \langle m_1, \ldots, m_k \rangle$, where $m_i = \langle s \mid V_w, seq(m_i), post(m_i) \rangle$ is a macro for a parent $w \in Pa(v)$ of $v$ such that $(s \mid V_w) \not\sim (x \mid V_w)$. Since $pre(m_i) = s \mid V_w$ holds in $s$, $m_i$ is applicable in $s$. Since the transitive reduction of the causal graph is an inverted tree, the parents of $v$ have no common ancestors, so the sets of variables $V_w$ are disjoint. As a consequence, the application of $m_1, \ldots, m_{i-1}$ does not change the values of variables in $V_w$, so $s \mid V_w$ still holds when $m_i$ is applied.

Assuming that the macro $m_i$ is well-defined, $(s \mid V_w) \oplus post(seq(m_i)) = post(m_i)$, and Compose$(v, s, x, O, G)$ only considers macros $m_i$ for $w$ such that $post(m_i) \sim (x \mid V_w)$. Since no other macro in $seq$ changes the values of variables in $V_w$, it holds that $(s \oplus post(seq)) \sim (x \mid V_w)$. If $(s \mid V_w) \sim (x \mid V_w)$ for a parent $w \in Pa(v)$ of $v$, $seq$ does not contain a macro for $w$, so $(s \oplus post(seq)) \mid V_w = s \mid V_w$, implying $(s \oplus post(seq)) \sim (x \mid V_w)$. Finally, no macro in $seq$ changes the value of $v$, so $(s \oplus post(seq))(v) = s(v)$. Since $x \sim (v = s(v))$ by definition and $(s \oplus post(seq)) \sim (x \mid V_w)$ for each $w \in Pa(v)$, it holds that $(s \oplus post(seq)) \sim x$. □

**Lemma A.2.** *For each $v \in V$ and each state $s$ for $V_v$, each macro $\langle pre(m), seq(m), post(m) \rangle$ returned by a call to* Solve$(v, s, Z, O, G)$ *is well-defined, and $post(m) \sim z$ for some $z \in Z$.*

*Proof.* The proof is a double induction on $v$ and the state-sequence pairs $(p, seq)$ visited during a call to Solve$(v, s, Z, O, G)$. In particular, we show that $seq$ is well-defined for $s$ and that $p = s \oplus post(seq)$. The first state-sequence pair removed from $Q$ on line 4 is $(s, \langle \rangle)$. Clearly, $\langle \rangle$ is well-defined for $s$ and $s = s \oplus post(\langle \rangle)$. Otherwise, assume that $seq$ is well-defined for $s$ and that $p = s \oplus post(seq)$. On line 8, a state-sequence pair $(p \oplus post(\langle seq2, a \rangle), \langle seq, seq2, a \rangle)$ is added to $Q$ for each action $a \in O$ such that $pre(a) \sim (v = p(v))$ and each sequence $seq2$ returned by Compose$(v, p, pre(a), O, G)$.

By induction, the macros returned by Solve$(w, s', Z', O', G)$ are well-defined for each $w \in Pa(v)$ and each state $s'$ for $V_w$. Lemma A.1 now implies that each sequence $seq2$ returned by Compose$(v, p, pre(a), O, G)$ is well-defined for $p$ and that $(p \oplus post(seq2)) \sim pre(a)$. In other words, $a$ is applicable in $p \oplus post(seq2)$, implying that $\langle seq2, a \rangle$ is well-defined for $p$. By assumption, $seq$ is well-defined for $s$ and $p = s \oplus post(seq)$, implying that $\langle seq, seq2, a \rangle$ is well-defined for $s$. If applied in $s$, $\langle seq, seq2, a \rangle$ results in the state $p \oplus post(\langle seq2, a \rangle)$, implying $p \oplus post(\langle seq2, a \rangle) = s \oplus \langle seq, seq2, a \rangle$.

On line 14, a macro $m = \langle s, \langle seq, seq2 \rangle, p \oplus post(seq2) \rangle$ is added to $M$ for each projected pre-condition $z \in Z$ such that $z \sim (v = p(v))$ and each sequence $seq2$ returned by Compose$(v, p, z, O, G)$. Lemma A.1 implies that $seq2$ is well-defined for $p$ and that $(p \oplus post(seq2)) \sim z$. Since $seq$ is well-defined for $s$ and $p = s \oplus post(seq)$, $\langle seq, seq2 \rangle$ is well-defined for $s$ and results in the state $p \oplus post(seq2)$ when applied in $s$. It follows that the macro $m$ is well-defined and that $post(m) = p \oplus post(seq2) \sim z$. □

We now proceed to prove Theorem 3.2. First note that MacroPlanner$(P)$ calls GetMacros$(v, init, A, G)$ to obtain a set of macros $M$, where $G = R$ is the transitive reduction of the causal graph of $P$ and $v$ is the root variable of $R$. Since each other variable is an ancestor of $v$, it follows that $V_v = V$ and $init \mid V_v = init$. The only projected pre-condition in $Z$ is that of the dummy action $a^*$, which equals $goal \mid V_v = goal$. Thus GetMacros$(v, init, A, G)$ calls Solve$(v, s, Z, O, G)$ with $s = init$ and $Z = \{goal\}$.

Lemma A.2 implies that each macro $m$ returned by Solve$(v, init, \{goal\}, O, G)$ is well-defined and that $post(m) \sim goal$. Since the pre-condition of each macro $m$ returned by Solve$(v, init, \{goal\}, O, G)$ is $init$ and $m$ is well-defined, the sequence $seq(m) = \langle a_1, \ldots, a_k \rangle$ is well-defined for $init$ and, when applied in $init$, results in the state $init \oplus post(seq(m)) = post(m) \sim goal$ which satisfies the goal state. Thus, $m$ solves $P$.

## Appendix B. Proof of Theorem 3.3

In this appendix we prove Theorem 3.3, which states that MacroPlanner$(P)$ returns an optimal solution to $P \in \mathbf{IR}$ if and only if one exists. First, we prove a series of lemmas which state that the macros returned by Solve represent the shortest solutions to the corresponding subproblems.

**Definition B.1.** *For each $v \in V$ and each pair of states $(s, t)$ for $V_v$, let $s \to t$ denote that there exists a sequence of actions in $A_v = \{a \in A : V_{post(a)} \subseteq V_v\}$ that, when applied in state $s$, results in state $t$.*

**Lemma B.2.** *For each $v \in V$, let $(p, seq)$ be a state-sequence pair visited during a call to* Solve$(v, s, Z, O, G)$. *Then for each $w \in Pa(v)$,* GetMacros$(w, init, A, G)$ *previously called* Solve$(w, p \mid V_w, Z', O', G)$, *where $Z'$ and $O'$ are the values of $Z$ and $O$ for $w$.*

*Proof.* GetMacros$(w, init, A, G)$ calls Solve$(w, s', Z', O', G)$ for each state $s'$ in the list $L$. If $p = init \mid V_v$ is the projected initial state for $V_v$, $p \mid V_w = init \mid V_w$ is added to $L$ on line 2 of GetMacros. Otherwise, the only actions in $O$ for changing the values of variables in $V_w$ are the macros generated by GetMacros$(w, init, A, G)$. The projection $p \mid V_w$ thus has to equal the post-condition of some such macro. Each distinct post-condition $post(m)$ of a macro $m$ for $w$ is added to $L$ on line 12 of GetMacros. $\square$

**Lemma B.3.** *For each $v \in V$, each pair of states $(s, u)$ for $V_v$ such that $s \to u$, and each projected pre-condition $z \in Z$ such that $u \sim z$,* Solve$(v, s, Z, O, G)$ *returns a macro $\langle s, seq, t \rangle$ such that $t \sim z$ and $t$ is on a shortest path from $s$ to $u$ with prefix $seq$.*

*Proof.* By induction on $v$. If $Pa(v) = \emptyset$, the set of actions is $O = A_v$, and $s \to u$ implies the existence of a sequence of actions in $A_v$ from $s$ to $u$. Since each projected pre-condition $z \in Z$ is non-empty, $u \sim z$ implies $u = z$. Since state-sequence pairs are visited in order of shortest sequence length, Solve$(v, s, Z, O, G)$ is guaranteed to return a macro $\langle s, seq, u \rangle$ such that $seq$ is a shortest path from $s$ to $u$. Clearly, $u$ is on a shortest path from $s$ to $u$, and $seq$ is a prefix of itself.

If $|Pa(v)| > 0$, we use induction on the state-sequence pairs $(p, seq)$ visited during the call to Solve$(v, s, Z, O, G)$ to prove the lemma. Namely, we prove that if $p \to u$, Solve$(v, s, Z, O, G)$ returns a macro $\langle s, \langle seq, seq2 \rangle, t \rangle$ such that $t \sim z$ and $t$ is on a shortest path from $p$ to $u$ with prefix $seq2$. The only exception to this rule is if there exists an action sequence from $s$ to $t$ that is shorter than $\langle seq, seq2 \rangle$ and does not pass through $p$.

The base case is given by $p(v) = u(v)$. For each $w \in Pa(v)$, $(p \mid V_w) \not\sim (z \mid V_w)$ implies $z \mid V_w \in Z'$, $u \sim z$ implies $(u \mid V_w) \sim (z \mid V_w)$, GetMacros$(w, init, A, G)$ previously called Solve$(w, p \mid V_w, Z', O', G)$ due to Lemma B.2, and $p \to u$ implies $(p \mid V_w) \to (u \mid V_w)$ by removing actions not in $A_w$. By induction, Solve$(w, p \mid V_w, Z', O', G)$ returned a macro $\langle p \mid V_w, seq', t' \rangle$ such that $t' \sim (z \mid V_w)$ and $t'$ is on a shortest path from $p \mid V_w$ to $u \mid V_w$ with prefix $seq'$. In particular, the macro $\langle p \mid V_w, seq', t' \rangle$ is now part of the set $O$.

In the iteration of SOLVE$(v, s, Z, O, G)$ for $(p, seq)$, COMPOSE$(v, p, z, O, G)$ is called on line 12 since $u \sim z$ and $p(v) = u(v)$ imply $z \sim (v = p(v))$. The resulting set $S$ contains a sequence $seq2$ which consists of the macros $\langle p \mid V_w, seq', t' \rangle$ for each $w \in Pa(v)$ such that $(p \mid V_w) \not\sim (z \mid V_w)$. On line 14 the macro $\langle s, \langle seq, seq2 \rangle, t \rangle$ is added to $M$, where $t = p \oplus post(seq2) \sim z$ due to Lemma A.1. Since $seq2$ does not change the value of $v$, since the subsets $V_w$ are disjoint, and since $t'$ is on a shortest path from $p \mid V_w$ to $u \mid V_w$ with prefix $seq'$ for each $w \in Pa(v)$, $t$ is on a shortest path from $p$ to $u$ with prefix $seq2$.

For $p(v) \neq u(v)$, any path from $p$ to $u$ has to include actions that change the value of $v$. Let $a \in A$ be the first such action on a shortest path from $p$ to $u$, and assume that $a$ is applied in state $r$, implying $p \to r$, $r \oplus post(a) \to u$, and $r \sim pre(a)$. For each $w \in Pa(v)$, $(p \mid V_w) \not\sim (pre(a) \mid V_w)$ implies $pre(a) \mid V_w \in Z'$, $p \to r$ implies $(p \mid V_w) \to (r \mid V_w)$, $r \sim pre(a)$ implies $(r \mid V_w) \sim (pre(a) \mid V_w)$, and GETMACROS$(w, init, A, G)$ called SOLVE$(w, p \mid V_w, Z', O', G)$ due to Lemma B.2. By induction, SOLVE$(w, p \mid V_w, Z', O', G)$ returned a macro $\langle p \mid V_w, seq', t' \rangle$ such that $t' \sim (pre(a) \mid V_w)$ and $t'$ is on a shortest path from $p \mid V_w$ to $r \mid V_w$ with prefix $seq'$. The macro $\langle p \mid V_w, seq', t' \rangle$ is now part of the set $O$.

In the iteration of SOLVE$(v, s, Z, O, G)$ for $(p, seq)$, COMPOSE$(v, p, pre(a), O, G)$ is called on line 6 since $pre(a) \sim (v = p(v))$ due to the fact that $a$ is the first action changing the value of $v$ on a shortest path from $p$ to $u$. The resulting set $S$ contains a sequence $seq2$ which consists of the macros $\langle p \mid V_w, seq', t' \rangle$ for each $w \in Pa(v)$ such that $(p \mid V_w) \not\sim (pre(a) \mid V_w)$. On line 8 the state-sequence pair $(q \oplus post(a), \langle seq, seq2, a \rangle)$ is added to $Q$, where $q = p \oplus post(seq2) \sim pre(a)$ due to Lemma A.1.

Since $seq2$ does not change the value of $v$ and $t'$ is on a shortest path from $p \mid V_w$ to $r \mid V_w$ with prefix $seq'$ for each $w \in Pa(v)$, $q$ is on a shortest path from $p$ to $r$ with prefix $seq2$. Consequently, there exists a shortest path $\langle seq2, seq3 \rangle$ from $p$ to $r$. This path does not change the value of $v$, else $a$ would not be applied in state $r$ on a shortest path from $p$ to $u$. Since $a$ is applicable in $q$ due to $q \sim pre(a)$, the sequence $\langle seq2, a, seq3 \rangle$ is a shortest path from $p$ to $r \oplus post(a)$. It follows that $q \oplus post(a)$ is on a shortest path from $p$ to $r \oplus post(a)$ with prefix $\langle seq2, a \rangle$. Thus $r \oplus post(a) \to u$ implies $q \oplus post(a) \to u$.

In a future iteration of SOLVE$(v, s, Z, O, G)$, the state-sequence pair removed from $Q$ on line 4 is $(q \oplus post(a), \langle seq, seq2, a \rangle)$. Since $q \oplus post(a) \to u$, by induction on state-sequence pairs, SOLVE$(v, s, Z, O, G)$ returns a macro $\langle s, \langle seq, seq2, a, seq4 \rangle, t \rangle$ such that $t \sim u$ and $t$ is on a shortest path from $q \oplus post(a)$ to $u$ with prefix $seq4$. Then there exists a shortest path $\langle seq4, seq5 \rangle$ from $q \oplus post(a)$ to $u$. Since $r \oplus post(a)$ is on a shortest path from $p$ to $u$ by assumption, and $q \oplus post(a)$ is on a shortest path from $p$ to $r \oplus post(a)$ with prefix $\langle seq2, a \rangle$, $\langle seq2, a, seq4, seq5 \rangle$ has to be a shortest path from $p$ to $u$. It follows that $t$ is on a shortest path from $p$ to $u$ with prefix $\langle seq2, a, seq4 \rangle$.

The proof now follows since the first iteration of SOLVE$(v, s, Z, O, G)$ is for $(s, \langle \rangle)$, implying that SOLVE$(v, s, Z, O, G)$ returns a macro $\langle s, \langle \langle \rangle, seq2 \rangle, t \rangle = \langle s, seq2, t \rangle$ such that $t \sim u$ and $t$ is on a shortest path from $s$ to $u$ with prefix $seq2$. In this case there can be no exception since any sequence from $s$ to $t$ passes through $s$. Note that if $v \notin V_z$, i.e., $z$ does not specify a value for $v$, the set of macros $M$ returned by SOLVE$(v, s, Z, O, G)$ contains macros to $z$ for each reachable value of $v$, including $u(v)$ since $u$ is reachable from $s$.  □

We now proceed to prove Theorem 3.3. Recall that the set of macros $M$ returned by MACROPLANNER$(P)$ is equal to the set of macros returned by SOLVE$(v, init, \{goal\}, O, G)$, where $G = R$ is the transitive reduction of the causal graph and $v$ is the root variable of $R$. An optimal plan solving $P$ is a sequence of actions in $A_v = A$ from $init$ to a state $u$ such that $u \sim goal$, implying $init \rightarrow u$. We can now apply Lemma B.3 to prove that $M$ contains a macro $m = \langle init, seq, t \rangle$ such that $t \sim goal$ and $t$ is on a shortest path from $s$ to $u$ with prefix $seq$. This is only possible if $seq$ is an optimal plan.

If there does not exist a plan solving $P$, we show that $M$ is empty by contradiction. Assume not. Then Theorem 3.2 implies that each macro in $M$ is a solution to $P$, which contradicts the fact that $P$ is unsolvable. Thus $M$ has to be empty, and as a consequence, MACROPLANNER($P$) returns "fail".

## Appendix C. Proof of Theorems 3.5 and 3.6

In this section we prove Theorems 3.5 and 3.6, which establish two subclasses of the class **IR** for which MACROPLANNER generates solutions in polynomial time.

To prove Theorem 3.5, note that for each variable $v \in V$ and each action $a$ with $v \in V_{post(a)}$, it holds that $V_{pre(a)} = V_v$. It follows that $v \in V_z$ for each projected pre-condition $z \in Z$. Let $Z_v^d = \{z \in Z : z(v) = d\}$ be the set of projected pre-conditions that specify the value $d$ for $v$. Any state in the domain transition graph that specifies the value $d$ for $v$ has to match either the pre-condition of an action in $A_v^d$ or a projected pre-condition in $Z_v^d$. Otherwise, the corresponding node would not be added to the graph by the algorithm. The only exception is the projected initial state $init \mid V_v$ in case $init(v) = d$.

Since each pre-condition is specified on each ancestor of $v$, only one state matches each pre-condition. The number of projected pre-conditions is bounded by the number of actions for descendants of $v$, so the number of nodes in the domain transition graph is $O(\sum_{d \in D(v)}(|A_v^d| + |Z_v^d|)) = O(|A| + |Z|) = O(|A|)$. From Lemma 3.4 it follows that the complexity of the algorithm is $O(|A| \sum_{v \in V} |A|^3) = O(|V||A|^4)$, proving the theorem.

Theorem 3.6 states that each action for changing the value of $v$ to $d \in D(v)$ has the same pre-condition on the parents of $v$. Since the pre-condition is undefined on all other ancestors, each projected pre-condition for $v$ is specified on $v$ alone. This implies that successor states of the first and second type always coincide, since matching a projected pre-condition only depends on the value of $v$.

We prove by induction that the domain transition graphs for $v$ contain at most one node for each value $d \in D(v)$. If $|Pa(v)| = 0$, the proof follows from the fact that the nodes are values of $D(v)$. If $|Pa(v)| > 0$, by induction the domain transition graphs of each parent $v' \in Pa(v)$ of $v$ has at most one node per value $d' \in D(v')$. This implies that each macro $m \in O$ with $post(m)(v') = d'$ has the same post-condition (else there would be at least two nodes for $v' = d'$).

For each value $d \in D(v)$, the algorithm generates a successor state $s$ of the first type with $s(v) = d$ by applying any action $a \in A_v^d$. Each such action $a$ has the same pre-condition on each parent $v' \in Pa(v)$. This implies that the pre-condition of $a$ is always satisfied in the same state, since any macro $m \in O$ with $post(m)(v') = pre(a)(v')$ has the same post-condition. Thus, any successor state $s$ with $s(v) = d$ is the same, no matter which action we use or which previous state we come from.

Note that for $d = init(v)$, the projected initial state $init \mid V_v$ could be different from the successor state $s_i$ discussed above. However, the theorem states that the pre-condition of actions with $post(a)(v) = init(v)$ on $Pa(v)$ equals $init \mid Pa(v)$. Since the domain transition graphs of each parent $v' \in Pa(v)$ contain at most one node for $init(v')$, this node has to correspond to the projected initial state $init \mid V_{v'}$. Thus, any successor state $s$ with $s(v) = init(v)$ equals the projected initial state $init \mid V_v$, since $pre(a) \mid Pa(v) = init \mid Pa(v)$ for each action $a \in A_v^d$, $d = init(v)$, and $post(m) = init \mid V_{v'}$ for each macro $m$ that satisfies the pre-condition of $a$. The proof of the theorem now follows from Lemma 3.4.

## Appendix D. Proof of Theorems 4.3 and 4.4

In this appendix we prove Theorems 4.3 and 4.4, which together state that RELAXED-PLANNER returns a well-defined optimal solution to $P \in \mathbf{RIR}$ if and only if one exists. We first show that Definition B.1 and Lemma B.2 apply to RELAXEDPLANNER. The notion of reachability in Definition B.1 refers to the set $A_v$ of actions such that $V_{post(a)} \subseteq V_v$, thus excluding actions that change the value of some successor of $v$.

Lemma B.2 states that if $(p, seq)$ is a state-sequence pair visited during a call to SOLVE$(v, s, Z, O, G)$ and $w \in Pa(v)$ is a parent of $v$ in $G$, GETMACROS$(w, init, A, G)$ previously called SOLVE$(w, p \mid V_w, Z', O', G)$, where $Z'$ and $O'$ are the values of $Z$ and $O$ for $w$. We show that this holds for RELAXEDMACROS as well, where $G$ is now the transitive reduction of the relaxed causal graph.

If $p$ equals the projected initial state $init \mid V_v$, $p \mid V_w = init \mid V_w$ is added to $L$ on line 2 of RELAXEDMACROS. Otherwise, $p$ is a successor state of the first type. In other words, it is reached by applying some action $a \in A'_v \subseteq A''_w$. SOLVE only uses macros for $w$ to satisfy the pre-condition of $a$ with respect to $V_w$. Consequently, $p \mid V_w$ equals the post-condition $post(m)$ of some such macro $m$, followed by the application of $a$. This is precisely the state $post(m) \oplus (post(a) \mid V_w)$ added to $L$ on line 15 of the modified RELAXEDMACROS.

Since the subroutines SOLVE and COMPOSE are the same as before, Lemmas A.2 and B.3 apply verbatim to RELAXEDPLANNER. We can now use the same reasoning as for MACROPLANNER to prove the two theorems.

## Appendix E. Proofs of Theorems 4.7, 4.8, and 4.9

In this appendix we prove Theorem 4.7, which states that the sequence $seq$ returned by REVERSIBLEPLANNER$(P)$ is well-defined for $init$ and satisfies $(init \oplus post(seq)) \sim goal$. We also prove Theorem 4.8, which states that REVERSIBLEPLANNER is complete for planning problems in $\mathbf{AR}$ with unary actions. Finally, we prove Theorem 4.9 regarding the complexity of REVERSIBLEPLANNER. We first prove lemmas similar to Lemmas A.1 and A.2.

**Lemma E.1.** *For each state $s$ and each partial state $x$, let $(seq, seq2)$ be the pair of sequences returned by* REVERSIBLECOMPOSE$(U, s, x, P, G)$. *If the macros returned by* REVERSIBLESOLVE *are well-defined, $seq$ is well-defined for $s$ and $s \oplus post(seq) = s \oplus x$. Furthermore, $seq2$ is well-defined for $s \oplus x$ and $post(seq2) = s \mid (V_x - U)$.*

*Proof.* Let $w_1, \ldots, w_k$ be a topological sort of the variables in $V_x$ such that $x(w_i) \neq s(w_i)$. The first sequence returned by REVERSIBLECOMPOSE$(U, s, x, P, G)$ is $seq = \langle m_k^1, \ldots, m_1^1 \rangle$, where $m_i^1 = \langle s \mid V_{w_i}, seq', (s \mid V_{w_i}) \oplus (w_i = x(w_i)) \rangle$ is the macro for changing the value of $w_i$ from $s(w_i)$ to $x(w_i)$. The fact that $w_i$ comes before $w_j$ in topological order implies $w_j \notin V_{w_i}$. Thus, after changing the values of $w_k, \ldots, w_{i+1}$, the pre-condition $s \mid V_{w_i}$ of the macro $m_l^1$ still holds, so the sequence $seq$ is well-defined for $s$. Furthermore, $seq$ changes the value of each variable $w \in V_x$ to $x(w)$, but leaves the values of all other variables unchanged, so $s \oplus post(seq) = s \oplus x$.

Let $u_1, \ldots, u_l$ be a topological sort of the variables in $W = \{w_1, \ldots, w_k\} - U$. The second sequence returned by REVERSIBLECOMPOSE$(U, s, x, P, G)$ is $seq2 = \langle m_1^2, \ldots, m_l^2 \rangle$, where $m_i^2 = \langle (s \mid V_{u_i}) \oplus (u_i = x(u_i)), seq', s \mid V_{u_i} \rangle$ is the macro for resetting the value of $u_i$ from $x(u_i)$ to $s(u_i)$. If $W$ contains ancestors of $u_i$, they come before $u_i$ in topological order. Hence their values are reset to $s$ prior to $u_i$, so the pre-condition $(s \mid V_{u_i}) \oplus (u_i = x(u_i))$ of the macro $m_i^2$ holds after the application of $m_1^2, \ldots, m_{i-1}^2$. Since $seq2$ resets the value of each variable $w \in V_x - U$ to $s(v)$, $post(seq2) = s \mid (V_x - U)$.

Note that if some ancestor $u'$ of $u_i$ were part of $U$, its value would not be reset to $s(u')$, and as a consequence, $seq2$ would not be well-defined. However, REVERSIBLESOLVE always calls REVERSIBLECOMPOSE with $U = V_{post(a)}$ and $x = pre(a)$ for some action $a$. Taken together, $u' \in V_{post(a)}$ and $u_i \in V_{pre(a)} - V_{post(a)}$ imply that the causal graph contains an edge $(u_i, u')$, which is true even if we consider the relaxed causal graph. Thus $u'$ could not be an ancestor of $u_i$ in an acyclic causal graph. □

**Lemma E.2.** *For each $v \in V$, each state $s$, and each value $d \in D(v)$, the macro returned by* REVERSIBLESOLVE$(v, s, d, A, G)$ *is well-defined.*

*Proof.* By induction on the state-sequence pairs $(p, seq)$ visited during a call to the subroutine REVERSIBLESOLVE$(v, s, d, A, G)$. In particular, we show that $seq$ is well-defined for $s$ and satisfies $s \oplus post(seq) = s \oplus p$. The base case is given by the first state-sequence pair $(s \mid W, \langle \rangle)$, whose sequence is clearly well-defined for $s$ and satisfies $s \oplus post(\langle \rangle) = s \oplus (s \mid W)$. Otherwise, assume that the statement holds for $(p, seq)$, and let $a$ be an action such that $pre(a) \sim (v = p(v))$.

Let $(seq2, seq3)$ be the result of REVERSIBLECOMPOSE$(V_{post(a)}, s \oplus p, pre(a), A, G)$. If $(seq2, seq3)$ does not equal ("fail", "fail"), Lemma E.1 states that $seq2$ is well-defined for $s \oplus p$ and that $s \oplus p \oplus post(seq2) = s \oplus p \oplus pre(a)$. This implies that $\langle seq, seq2, a \rangle$ is well-defined for $s$ and that $s \oplus post(\langle seq, seq2, a \rangle) = s \oplus p \oplus pre(a) \oplus post(a)$. Lemma E.1 also states that $seq3$ is well-defined for $s \oplus p \oplus pre(a)$ and that $post(seq3) = (s \oplus p) \mid (V_{pre(a)} - V_{post(a)})$. Since $seq3$ only changes the values of variables in $V_{pre(a)} - V_{post(a)}$, it is still well-defined in the state $s \oplus p \oplus post(a)$ that results from applying $a$. Thus $\langle seq, seq2, a, seq3 \rangle$ is well-defined for $s$ and $s \oplus post(\langle seq, seq2, a, seq3 \rangle) = s \oplus p \oplus post(a)$. It is easy to show that composition is commutative, implying $s \oplus p \oplus post(a) = s \oplus (p \oplus post(a))$. Thus the statement holds for the new state-sequence pair $(p \oplus post(a), \langle seq, seq2, a, seq3 \rangle)$ inserted into $L$ on line 12 of REVERSIBLESOLVE.

The proof now follows by induction on variables $v \in V$. If $v$ has no ancestors in the causal graph, any call to REVERSIBLECOMPOSE$(V_{post(a)}, s \oplus p, pre(a), A, G)$ returns a pair of empty sequences, since $pre(a) \sim (v = p(v))$ and the set $V_{pre(a)} - V_{post(a)}$ is empty. Thus the macro returned by REVERSIBLESOLVE is well-defined. Otherwise, by hypothesis of induction the macros generated by REVERSIBLESOLVE for ancestors of $v$ are well-defined. Thus REVERSIBLECOMPOSE$(V_{post(a)}, s \oplus p, pre(a), A, G)$ returns a well-defined pair of sequences due to Lemma E.1 and consequently, the macros generated by ReversibleSolve for $v$ are also well-defined. □

The proof of Theorem 4.7 now follows by a straightforward application of Lemma E.1 for the call ReversibleCompose($\emptyset, init, goal, A, G$) made by ReversiblePlanner, taking advantage of Lemma E.2 to ensure that the macros returned by ReversibleSolve are well-defined.

To prove Theorem 4.8, we first prove a lemma regarding the completeness of ReversibleSolve for unary actions.

**Lemma E.3.** *For each $v \in V$, each state $s$, and each value $d \in D(v)$, if there exists a state $u$ such that $u(v) = d$ and $s \to u$, ReversibleSolve$(v, s, d, A, G)$ returns a valid macro different from "fail".*

*Proof.* By a double induction on variables $v \in V$ and the state-sequence pairs $(p, seq)$ visited during a call to ReversibleSolve$(v, s, p, A, G)$. In particular, we show that if $s \oplus p \to u$, ReversibleSolve$(v, s, d, A, G)$ returns a valid macro. The base case is given by $p(v) = d$, implying $p = (s \mid W) \oplus (v = d)$ since $W = \emptyset$ for unary actions. In this case, ReversibleSolve returns a macro on line 7. Else the path from $s$ to $u$ has to contain actions for changing the values of $v$. Let $a$ be the first such action, implying $s \to t$ for some state $t$ such that $t \sim pre(a)$.

By induction, for each $w \in V_{pre(a)}$, ReversibleSolve$(w, s, pre(a)(w), A, G)$ returns a valid macro. The fact that $w$ is reversible implies that there exists a state $u'$ such that $u'(w) = s(w)$ and $s \oplus (w = pre(a)(w)) \to u'$. As a consequence, by induction ReversibleSolve$(w, s \oplus (w = pre(a)(w)), s(w), A, G)$ also returns a valid macro. Thus the pair of sequences $(seq2, seq3)$ returned by ReversibleCompose$(V_{post(a)}, s \oplus p, pre(a), A, G)$ is well-defined and different from ("fail", "fail"). Consequently, ReversibleSolve inserts the state-sequence pair $(p \oplus post(a)(v), \langle seq, seq2, a, seq3 \rangle)$ into the list $L$ on line 12.

The sequence $\langle seq, seq2, a, seq3 \rangle$ generated by ReversibleSolve results in the state $s \oplus (v = post(a)(v))$, which may be different from the state on the given path from $s$ to $u$ after applying action $a$. However, since $v$ is reversible, that other state is reachable from $s \oplus (v = post(a)(v))$, implying $s \oplus (v = post(a)(v)) \to u$. By induction on state-sequence pairs, ReversibleSolve$(v, s, d, A, G)$ returns the corresponding macro. □

The proof of Theorem 4.8 now follows from the fact that in the call to the subroutine ReversibleCompose($\emptyset, init, goal, A, G$), if each value in the goal state can be satisfied starting in $init$, Lemma E.3 implies that ReversibleSolve is guaranteed to return a macro that satisfies the goal state for each variable in $V_{goal}$. Thus, due to Lemma E.1, ReversibleCompose will successfully return a pair of well-defined sequences $(seq, seq2)$ with the property $init \oplus post(seq) = init \oplus goal$, so $seq$ is a solution to $P$. If some value in the goal state is unreachable from $init$, ReversibleSolve does not return a corresponding macro, so ReversibleCompose (and ReversiblePlanner) return "fail".

To prove Theorem 4.9, we first prove a lemma regarding the values of variables in $V_v - W$:

**Lemma E.4.** *For each $v \in V$ and any call to ReversibleSolve$(v, s, d, A, G)$, it holds that $s \mid (V_v - W) = init \mid (V_v - W)$.*

*Proof.* The lemma states that variables in $V_v - W$ always take on their initial values in any call to REVERSIBLESOLVE. First note that REVERSIBLEPLANNER calls REVERSIBLECOMPOSE with $s = init$. In turn, REVERSIBLECOMPOSE makes two calls to REVERSIBLESOLVE for each variable $w \in V_x$. In these calls, the only variable that may have a different value from that in $s$ is $w$. However, for any call to REVERSIBLESOLVE for $w$, it holds that $w \in W$ (else there is no action for changing the value of $w$, so we can remove it from the problem). Thus the lemma holds for the call to REVERSIBLECOMPOSE by REVERSIBLEPLANNER.

Now assume that the lemma holds for a call to REVERSIBLESOLVE$(v, s, d, A, G)$. In a call to REVERSIBLECOMPOSE, the value of $s$ is $s \oplus p$, where $p$ is a partial states on the variables in $W$. Note that because of the definition of $W$, if $w \in W$ for $v$, $w \in W$ for any ancestor $u$ of $v$ such that $w \in V_u$. Thus if the value of $w$ is different from $init(w)$ in a call to REVERSIBLECOMPOSE, $w \in W$ in any subsequent call to REVERSIBLESOLVE. Since all variables not in $W$ take on their initial values by assumption, they take on their initial values in subsequent calls to REVERSIBLESOLVE as well. □

Because of Lemma E.4, the number of distinct choices of $s$ in calls to REVERSIBLESOLVE for a variable $v$ is $O(\mathcal{D}^k)$, where $k = |W|$. Since the number of distinct choices of $d$ is $O(\mathcal{D})$, the total number of calls to REVERSIBLESOLVE for $v$ is $O(\mathcal{D}^{k+1})$. Each call to REVERSIBLESOLVE for a variable $v \in V$ is a breadth-first search in a graph with $O(\mathcal{D}^k)$ nodes. For each action $a \in A'$, there is at most one edge for each node, so the complexity of breadth-first search is $O(\mathcal{D}^k(1 + |A'|)) = O(\mathcal{D}^k|A'|)$, and the total complexity of calls to REVERSIBLESOLVE for $v$ is $O(\mathcal{D}^{2k+1}|A'|)$. Since the sets $A'$ are distinct, the total complexity of calls to REVERSIBLESOLVE is $O(\sum_{v \in V} \mathcal{D}^{2k+1}|A'|) = O(\mathcal{D}^{2k+1}\sum_{v \in V}|A'|) = O(\mathcal{D}^{2k+1}|A|)$.

REVERSIBLECOMPOSE is called at most $\mathcal{D}^k|A|$ times, once for each edge in one of the graphs traversed by REVERSIBLESOLVE. The worst-case complexity of REVERSIBLECOMPOSE is linear in the number of variables in $V_x$, or $O(|V|)$. The total complexity of REVERSIBLECOMPOSE is thus $O(\mathcal{D}^k|A||V|)$, so the total complexity of REVERSIBLEPLANNER is $O(\mathcal{D}^{2k+1}|A| + \mathcal{D}^k|A||V|) = O(\mathcal{D}^k|A|(\mathcal{D}^{k+1} + |V|))$.